\documentclass{article}
\usepackage[T1]{fontenc}
\usepackage{iclr2027_conference,times}

\usepackage[utf8]{inputenc}
\usepackage{graphicx}
\usepackage{amsmath,amssymb,amsfonts}
\usepackage{algorithm}
\usepackage{algorithmic}
\usepackage{booktabs}
\usepackage{multirow}
\usepackage{array}
\usepackage{longtable}
\usepackage{listings}
\usepackage{placeins}
\usepackage{xcolor}
\usepackage{url}
\usepackage[colorlinks=true,allcolors=blue]{hyperref}
\hypersetup{
  pdftitle={FARL: A Fully Automated Pipeline of RL Interface and Solver Design},
  pdfauthor={Tianyu Jin, Liuyu Xiang, Chaowen Yang, Xingyang Chang, Pei Pei Li, Zhaofeng He},
  pdfsubject={Research preprint},
  pdfkeywords={automated reinforcement learning, interface and solver design}
}

\newcommand{\stageomr}{\ensuremath{\mathcal{S}}}
\newcommand{\finalomr}{\ensuremath{\mathcal{F}}}
\title{FARL: A Fully Automated Pipeline of\\
RL Interface and Solver Design}
\author{
\textbf{Tianyu Jin \quad Liuyu Xiang\thanks{Corresponding author: \texttt{xiangliuyu@bupt.edu.cn}.} \quad Chaowen Yang} \\
\textbf{Xingyang Chang \quad Pei Pei Li \quad Zhaofeng He} \\
{\normalfont Beijing University of Posts and Telecommunications, Beijing, China}
}
\date{}

\iclrfinalcopy

\newcommand{\handSuccess}{97.11}
\newcommand{\handSD}{2.99}
\newcommand{\handSuccessRounded}{97.1}
\newcommand{\handTokens}{2.58}
\newcommand{\handSteps}{408.4}
\newcommand{\handReach}{100.0}
\newcommand{\handLift}{99.0}
\newcommand{\handTransport}{97.8}

\begin{document}
\maketitle
\lhead{Preprint}
\raggedbottom
\suppressfloats[t]

\begin{abstract}
Existing reinforcement learning (RL) largely focuses on optimizing learning algorithms within a pre-defined task formulation, while applying RL to new tasks still relies heavily on human expertise to design the learning pipeline: what an agent observes, how it is rewarded, which algorithm it uses, and how training is configured. Automating these choices is challenging because they are tightly coupled. We introduce FARL (Fully Automated Reinforcement Learning), a framework that automates RL design by jointly searching observation and reward programs, learning algorithms, and training settings. FARL uses training statistics and observed behavior to diagnose failures, propose changes, and test whether they improve learning. Its search preserves diverse candidates and the controls needed for these tests, using a staged budget to screen designs before investing in longer training. Experimental memory records each finding together with its conditions, so an unsuccessful component can be reconsidered in a different combination. Across expert-withheld and expert-available settings, FARL outperforms the evaluated automated baselines on challenging navigation, locomotion, and manipulation tasks. On Adroit Hand Relocate, FARL achieves \handSuccessRounded\% success at 5M training steps per policy, compared with 0\% for Eureka and LiMEN and 1.0\% for native Proximal Policy Optimization (PPO) under the same policy training budget.

\end{abstract}

\section{Introduction}\label{sec:intro}

Applying reinforcement learning to a new task requires decisions about
observations, rewards, learning algorithms, and training settings. These
choices still depend heavily on task-specific expertise. Automated reinforcement
learning (AutoRL) aims to reduce this dependence by designing the learning
setup as well as training the policy~\citep{autorl_survey2022}.

\begin{figure}[t]
\centering
\includegraphics[width=\textwidth]{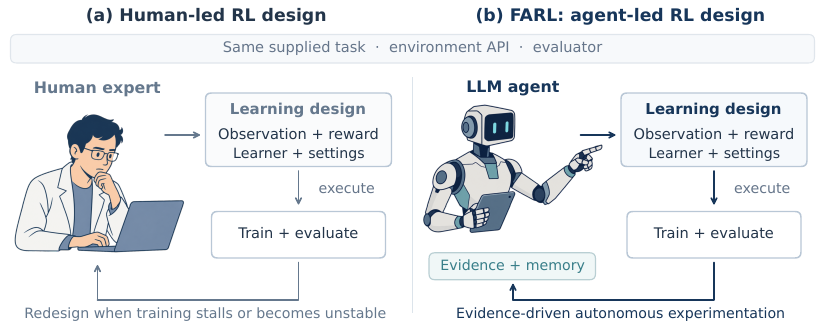}
\caption{Automating the RL design loop. (a) An expert revisits the learning
setup when training stalls or becomes unstable. (b) FARL's LLM agent designs
and directs experiments using behavioral evidence and scientific memory.
Both receive the same task interface and evaluator.}
\label{fig:overview}
\end{figure}

The difficulty becomes apparent when training fails. A low task score
does not reveal whether the reward encourages the wrong behavior, the
observations omit useful information, or the learner cannot exploit the
interface. Changing one component may alter the value of the others.
Early scores can also be misleading: a slowly learning design may
eventually outperform an early leader. Training every combination to
completion is too expensive, yet rejecting weak early candidates can
remove useful alternatives. An automated designer must therefore decide
what to change and how much training is needed to assess that change.

Existing methods automate several of these choices. Eureka generates
reward programs, while LiMEN searches observations and
rewards~\citep{eureka2024,limen2026}. Other systems use behavioral feedback
to revise rewards and training settings, or automate task formulation
followed by algorithm optimization~\citep{rda2026,prompt2policy2026,agent22025}.
We study how to revisit these choices together when failure leaves
several plausible explanations.

We introduce FARL, which designs an RL learning setup from a task's
application programming interface (API) and evaluator
(Figure~\ref{fig:overview}). \textbf{Interface--Solver Co-Design}
keeps observations, rewards, learners, and training settings open to
revision. \textbf{Falsifiable Diagnosis} uses training and trajectory
evidence to propose changes, predicts their effects, and tests them by
comparing the resulting policies.

\textbf{Diversity-Preserving Search} keeps promising candidates and the
controls needed for these comparisons. Short trials screen a broad set
of designs before selected candidates receive longer training. Scientific
memory records each result with its test conditions, so a component that
failed in one combination can be reconsidered in another. This allows
the search to correct earlier decisions as new results become available.

Our contributions are as follows.

\noindent\textbf{(1)} We introduce FARL, a framework for fully automated
RL interface and solver design. It combines joint design, falsifiable
diagnosis, and diversity-preserving search in a closed loop, using
experimental findings to revise candidates and allocate training.

\noindent\textbf{(2)} We address the difficulty of diagnosing failures
across coupled design choices under limited training budgets. FARL tests
specific changes while retaining alternative configurations and records
the conditions of each result. The search can then revisit unsuccessful
combinations and continue training candidates whose progress emerges later.

\noindent\textbf{(3)} Experiments show that FARL discovers effective
designs without native expert guidance and improves difficult
manipulation: it solves DoorKey where Eureka and LiMEN fail, reaches
3.3 times Eureka's Ant displacement at 5M steps, and achieves
\handSuccessRounded\% Hand relocation success versus 0\% for both automated
baselines at 5M steps per policy. Search histories,
controlled comparisons, and ablations examine how its design and
selection decisions contribute to these results (Section~\ref{sec:experiments}).

\section{Comparison to Prior Work}\label{sec:related}

Eureka generates reward programs~\citep{eureka2024}. LiMEN searches
observations and rewards using a quality-diversity archive and increasing
training budgets, with a fixed learner and training
configuration~\citep{limen2026}. Agent$^2$ first models the MDP and then
optimizes the algorithm, network, and hyperparameters using training
feedback~\citep{agent22025}. FARL keeps both the interface and solver open
to revision, so the same observation and reward programs can be tested
with different learners and training settings throughout search.

MAP-Elites maintains diverse solutions, while successive-halving methods
devote more training to promising candidates~\citep{mouret2015mapelites,li2018hyperband,li2020asha}.
FARL also retains candidates whose behavior is still improving or whose
comparison can test a diagnosis. Their components remain available for
new combinations even if the original design is unsuccessful.

Reward Design Agent (RDA) uses visual subtask diagnosis to revise
rewards~\citep{rda2026}. Prompt-to-Policy combines behavioral feedback,
training analysis, and previous lessons to revise rewards and
hyperparameters~\citep{prompt2policy2026}. RHyVE tests reward hypotheses
through short training forks from shared checkpoints and uses their
phase-dependent effects to guide reward deployment~\citep{rhyve2026}.
FARL tests changes to the interface or solver, specifying what changes,
what stays fixed, and what behavioral improvement is expected. It records
the actual outcome and test conditions for use in later design and
training decisions. Appendix~\ref{app:prior-work} gives the broader comparison.

\section{Problem and Information Settings}\label{sec:problem}
FARL receives a task description, an environment with documented observation
and action fields, a task evaluator $\finalomr$, and resource limits and
stopping rules $B$. It constructs a learning design
\[
h=(\underbrace{\phi,r}_{\text{learning interface}},\underbrace{L,\lambda,e,b}_{\text{solver configuration}}),
\]
where $\phi$ maps observations to policy inputs, $r$ is a reward program,
$L$ a supported learner, $\lambda$ its hyperparameters, $e$ its exploration
procedure, and $b$ the number of training steps per policy. Exploration may
include temporary assistance from a previously trained policy
(Section~\ref{sec:breadth}). For an evaluation episode $\tau$, $\finalomr(\tau)$
is the task outcome; $\finalomr_s(h)$ averages this outcome over episodes
for independently initialized training run $s$. Evaluation starts from the
original reset distribution with the trained policy acting alone. FARL uses these measurements to select designs and decide which to train further.

The evaluator and optional milestone rules are fixed before search and
applied to every method.
Milestones measure intermediate behaviors, such as picking up a key or
lifting an object. They let FARL identify where behavior breaks down and test whether a
design change addresses that failure.
Appendix~\ref{app:hand-champion-cohorts} gives their exact definitions.

\paragraph{Information settings and their purpose.}
Mature benchmarks package a task together with expert-designed rewards and
learning interfaces. We vary access to this prior design work while keeping
the task dynamics, reset distribution, and evaluator fixed.
\emph{Expert-withheld} (EW) removes native rewards and additional engineered
or privileged inputs from candidate training, retaining the public API,
observations, actions, goal, and descriptions of the available fields. This setting tests whether a method can design a learning setup without those expert choices.
\emph{Expert-available} (EA) allows native rewards, observations, and other
permitted task information, asking whether automated design can improve on
the available starting point. We use \emph{stress test} (ST) for the difficult
EA case where native designs still perform poorly: the challenge is to make
existing information useful for learning. Adroit Hand Relocate serves this
role. Task-specific interfaces and access rules are detailed in
Appendix~\ref{app:cohort-interfaces}.

\section{FARL: A Scientific Design--Train--Diagnose Loop}\label{sec:method}

\begin{figure}[t]
\centering
\includegraphics[width=\textwidth]{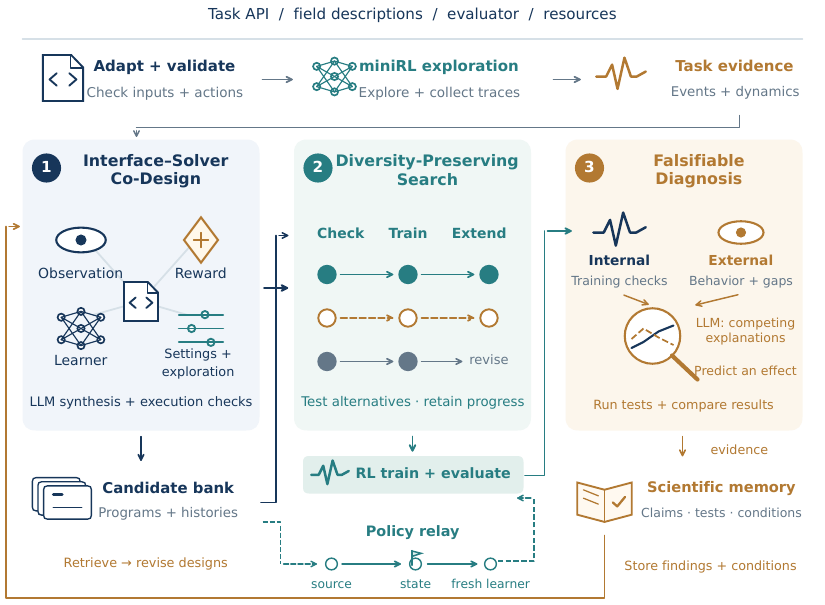}

\caption{FARL's architecture. An adapter exposes the task API, and miniRL
collects initial experience. (1) LLMs propose observations, rewards, and
solver configurations. (2) Search screens diverse candidates at short
budgets and trains selected designs and controls for longer. (3) Diagnosis
uses training statistics and behavior to propose the next experiment.
Programs enter the candidate bank; findings and test conditions enter
scientific memory. A trained policy may help a new learner reach an
intermediate state during training. Final evaluation uses the new policy
alone from the original resets.}

\label{fig:framework}
\end{figure}

FARL repeatedly proposes learning designs, trains policies, and uses the
results to decide what to test next (Figure~\ref{fig:framework}). A
\emph{candidate bank} stores reusable programs and configurations.
\emph{Scientific memory} stores the explanations tested, the outcomes,
and the settings under which they were obtained.

\subsection{Interface--Solver Co-Design}\label{sec:llm}
\paragraph{Collect initial task experience.}

FARL obtains a reset/step API from a registered environment constructor or
a tested adapter. Field definitions and access permissions specify what
candidate programs may use. Random actions and a short exploratory
learner, miniRL, collect trajectories that reveal which effects the agent
can produce and which events are difficult to reach. miniRL uses permitted
task feedback or a novelty reward (Appendix~\ref{app:adapter-bootstrap}).

\paragraph{From task evidence to an executable learning design.}

The generator receives the task goal, permitted inputs, trajectories,
candidate results, and relevant findings from earlier experiments. It
outputs an observation program $\phi$, a reward program $r$, and compatible
solver settings $(L,\lambda,e,b)$, together forming a learning design $h$.
Observation programs select coordinates or compute features such as
differences and distances. Reward programs use transitions, actions, and
episode-local memory to encourage progress or penalize setbacks.
Proposals can explain the learning difficulty they address and predict
how behavior should change. Separate policy and reward interfaces allow
the same reward to be tested with different observation maps.

\paragraph{Make solver choice part of that same design.}
Solver proposals specify the learner, network, hyperparameters, and
exploration procedure. Exploration choices include entropy settings,
random-action warmup, and training-only novelty rewards; hyperparameter
search follows the requirements of the chosen learner. The generator
can revise a component, combine earlier proposals, or introduce a new
design. Programs are checked for compatibility and execution errors
before training. A reward that fails with one solver remains available
for testing with another (Appendices~\ref{app:typed-programs}
and \ref{app:operational-choices}).

\subsection{Diversity-Preserving Search}\label{sec:breadth}
\paragraph{Cover alternative ways to learn.}
The bank defines a space of compatible observation--reward--learner--exploration
combinations,
\[
\mathcal Q=\{(\phi,r,L,e):\phi\in\mathcal C_\phi,\ r\in\mathcal C_r,
L\in\mathcal C_L,\ e\in\mathcal C_e,
\operatorname{compatible}(\phi,r,L,e)\}.
\]
Here $\mathcal C_\phi,\mathcal C_r,\mathcal C_L,\mathcal C_e$ are the
observation, reward, learner, and exploration components currently in the bank.

FARL first selects candidates needed for diagnostic comparisons, including
their controls. It then uses the remaining budget to cover more components
and component pairs. This avoids exhaustively training the full space.
Candidates are grouped by learner, observation type, reward construction,
and exploration procedure using their definitions and proposal labels.

\paragraph{Give promising designs more training.}
FARL follows a successive-halving-style budget
schedule~\citep{li2018hyperband,li2020asha}. The first short trial checks
whether a design runs correctly; zero task success at this stage is not
enough to reject it. Later trials use longer training and, where specified,
more seeds. A candidate can be retained for three reasons:
\emph{competitiveness}, indicated by strong task results;
\emph{learning potential}, indicated by improving milestones or variation
across runs; and \emph{experimental value}, such as serving as a control.
When progress stalls, FARL can revise the design, repeat an uncertain
comparison, or train another candidate. The fixed evaluator allows
comparisons across reward programs, whose training returns need not be
comparable (Appendix~\ref{app:retention-tests}). The bank also preserves
programs from unsuccessful combinations so they can be tested with new
learners or training settings.

\paragraph{Use learned behaviors to explore later task stages.}
FARL can freeze a policy that reaches an intermediate goal and let a
fresh learner take control from there. This \emph{policy relay} lets a
lifting policy, for example, help a new learner practice transport.
The new learner has its own parameters and optimizer. Assisted episodes
start from ordinary resets and are gradually replaced by unassisted
episodes. Final evaluation uses the new policy alone from the original resets
(Appendix~\ref{app:frontier-exploration}).

\subsection{Falsifiable Diagnosis}\label{sec:controller}\label{sec:execution}
\paragraph{Examine training and behavior together.}

The diagnostic LLM examines learning curves, optimizer statistics, and
whether higher reward corresponds to better task performance. It checks
which milestones the policy reaches, where progress stops, and whether
individual trajectories contradict an apparent improvement. Execution
and optimizer checks help separate implementation faults from learning
difficulties. These observations guide the choice between revising a
design, extending training, and running a controlled comparison. Task
performance is measured by the fixed evaluator~\citep{pan2022effects,skalse2022reward}.

\paragraph{State what a proposed change should improve.}
A diagnosis proposes an explanation for a failure and predicts the effect
of a change. A policy that touches an object but rarely moves it toward
the target, for example, may be rewarded for approach without maintaining
contact. Alternatively, its observations may make the target relation
difficult to learn. The first explanation suggests changing the reward
and testing whether the policy holds the object longer; the second
suggests changing the observations and testing for more target-directed
movement. FARL chooses the training budget and measurements needed for
the test. It can change several components when searching for better
performance, or hold the others fixed to examine a specific explanation.

\paragraph{Test the prediction and retain the result.}
For each comparison, the controller specifies the changed and fixed
components, the predicted effect, the seeds, and the training budget.
It checks that the intended programs and settings were executed before
interpreting the outcome. An inconclusive comparison may lead to more
training, another seed, or a different test.

Scientific memory records the explanation, experiment, outcome, and test
conditions. When proposing later designs, FARL retrieves relevant results,
including those that contradicted earlier predictions. A failure with one
learner or budget can therefore inform a new comparison without ruling
out the component in every setting. The candidate bank keeps those
components available. Appendix~\ref{app:operational-choices} describes
the controller and memory rules; Algorithm~\ref{alg:aspire} summarizes
the procedure.

\FloatBarrier
\section{Experiments}\label{sec:experiments}
We test whether FARL can construct effective designs when expert guidance
is withheld and improve on native configurations that struggle to learn.
We use learning curves, controlled comparisons, and ablations to examine how FARL finds and improves these designs.
\paragraph{Tasks and implementation.}
We evaluate $5\times5$ DoorKey navigation, MuJoCo Ant locomotion,
Meta-World Door-Open as a manipulation calibration task, and Adroit Hand
Relocate as a stress test~\citep{yu2020metaworld,chevalier2023minigrid,todorov2012mujoco,
rajeswaran2018dapg,gymnasiumrobotics2024}.
Appendices~\ref{app:cohort-interfaces} and \ref{app:design-comparison} give the
interfaces and compare native designs with the programs and solver settings
selected by each method.
Inner learners use Stable-Baselines3 (SB3)~\citep{raffin2021sb3}; the EW
searches use gateway model alias \texttt{gpt-5.4}. Appendix~\ref{app:search-cost}
gives the LLM settings and recorded expenditure.

\paragraph{Measurement and computation.}
DoorKey reports task completion; Hand requires the native relocation-success
predicate for five consecutive steps. Ant reports $\finalomr(\tau)=x_T-x_0$ in meters, the signed change in the
root position along the forward axis, where $T$ is environment termination
or 1{,}000 control steps, whichever comes first. Like Eureka and LiMEN, we evaluate locomotion by forward progress~\citep{eureka2024,limen2026}, using the units and horizon defined here.
Each training seed initializes an independent policy run. We average its
evaluation episodes, then report the mean and sample standard deviation
(SD) across runs~\citep{agarwal2021precipice}. Table~\ref{tab:mainresults} reports outcomes and costs; Appendices~\ref{app:cohort-details} and \ref{app:search-cost} detail evaluation and expenditure.
Policy training budgets describe the selected designs; total search costs
include the experiments used to find them. DoorKey EW uses 100k screening and 300k comparisons before a separate 1M evaluation; Ant and Hand budgets are detailed below.

\begin{table}[!t]
\centering
\caption{Task outcomes (mean $\pm$ SD) and costs (tokens / training steps,
both in M). Ant uses meters, Hand percent, other tasks completion rates.
Ant entries use 10M for FARL/LiMEN and 5M for Eureka/native PPO.
All Hand policies use 5M steps. Ant and Hand costs cover one complete search.
Bold marks the highest outcome or lowest cost in each column, including ties.
`/` means inapplicable.
$^*$Estimated tokens; $^\dagger$trimmed statistic; $^\ddagger$episode SD.
Policy horizons and accounting: Appendices~\ref{app:cohort-details}
and \ref{app:search-cost}.}
\label{tab:mainresults}
\setlength{\tabcolsep}{2pt}
{\footnotesize\renewcommand{\arraystretch}{1.1}
\begin{tabular*}{\textwidth}{@{\extracolsep{\fill}}lrrrrrr@{}}
\toprule
 & \multicolumn{2}{c}{DoorKey} & \multicolumn{2}{c}{Ant (m)} & Door & Hand (\%) \\
\cmidrule(lr){2-3}\cmidrule(lr){4-5}
Method & EW & EA & EW & EA & EW & ST \\
\midrule
FARL & $\boldsymbol{1.00\pm0.00}$ & $\boldsymbol{1.00\pm0.00}$ & $\boldsymbol{345.8\pm14.2}$ & $\boldsymbol{281.1\pm74.4}$ & $\boldsymbol{1.00}$ & $\boldsymbol{\handSuccess\pm\handSD}$ \\
Eureka & $0.00\pm0.00$ & $0.92\pm0.06$ & $94.1\pm50.9$ & $110.8\pm39.5$ & $\boldsymbol{1.00}$ & $0.00\pm0.00$ \\
LiMEN & $0.00\pm0.00^\ddagger$ & $\boldsymbol{1.00\pm0.00}$ & $89.9\pm20.4$ & $144.2\pm4.7^\dagger$ & $\boldsymbol{1.00}$ & $0.00\pm0.00$ \\
\midrule
Native design & / & $\boldsymbol{1.00\pm0.00}$ & / & $81.9\pm19.8$ & / & $1.00\pm1.00$ \\
\bottomrule
\end{tabular*}
\par\vspace{3pt}
\begin{tabular*}{\textwidth}{@{\extracolsep{\fill}}lrrrr@{}}
\toprule
Costs (M) & DoorKey EW & Ant EW & Door EW & Hand ST \\
\midrule
FARL & $2.834\,/\,82.3$ & $0.2^*\,/\,1{,}900.7$ & $0.874\,/\,\mathbf{9.1}$ & $\handTokens^*\,/\,\handSteps$ \\
Eureka & $\mathbf{0.249}\,/\,85$ & $\mathbf{0.152}\,/\,\mathbf{425}$ & $\mathbf{0.015}\,/\,82.5$ & $\mathbf{0.284}\,/\,110.4$ \\
LiMEN & $1.095\,/\,\mathbf{30}$ & $0.429\,/\,900$ & $0.577\,/\,195$ & $0.607\,/\,\mathbf{55.0}$ \\
\bottomrule
\end{tabular*}}
\end{table}

\subsection{Discovering Effective Learning Designs}\label{sec:results}\label{sec:cases}
\paragraph{Door: calibrating design without expert guidance.}
All three automated methods achieve 100\% completion on Door under EW
(Table~\ref{tab:mainresults}), confirming that existing methods can construct
successful designs without native expert rewards on this calibration task.

\begin{figure}[t]
\centering
\includegraphics[width=\textwidth]{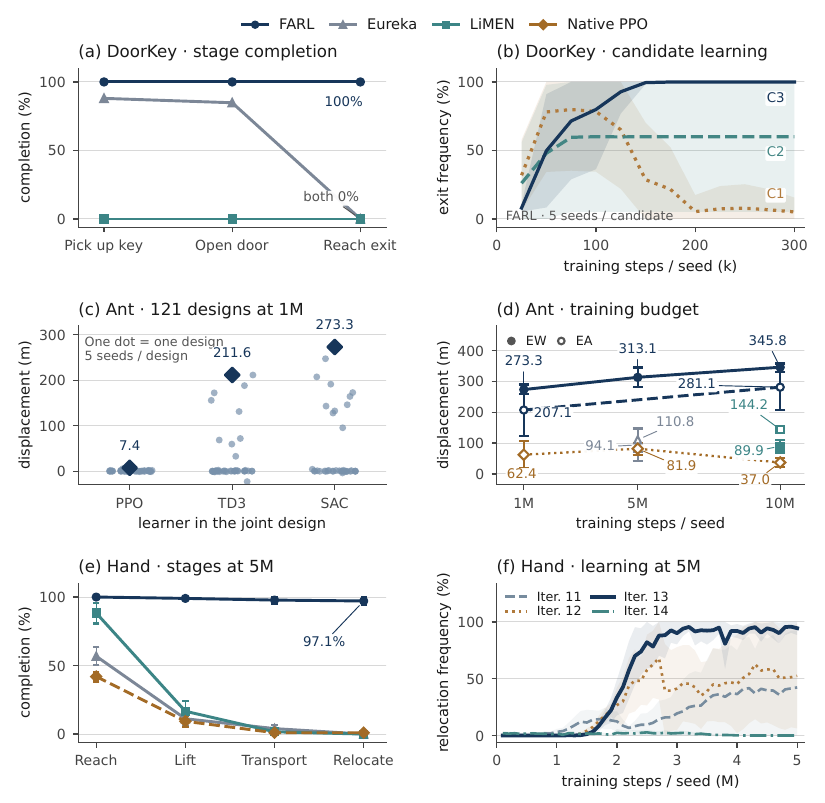}
\caption{Outcomes, design diversity, and learning dynamics.
(a) DoorKey stage completion. (b) Curve 1 (C1), Curve 2 (C2), and Curve 3 (C3)
track three FARL candidates at 300k: C1/C2 use
PPO, C3 uses A2C; lines are five-seed mean exit frequencies in 25k-step windows.
(c) All 121 Ant designs at 1M: dots are five-seed means; diamonds mark each
learner's best design. (d) Ant outcomes by budget; filled/open markers denote
EW/EA. (e) Hand stage completion at 5M steps per policy. (f) The 111-dimensional SAC design across
Hand iterations 11--14; lines average relocation frequencies across three training
runs in 100k-step windows.
Bars and bands show seed SD, clipped to $[0,100]\%$ for completion.
Appendices~\ref{app:learning-curves} and \ref{app:hand-breakthrough} give
candidate identities and final-evaluation protocols.}
\label{fig:funnel}
\end{figure}

\paragraph{DoorKey: finding a design that completes the task.}
Table~\ref{tab:mainresults} shows 100\% EW completion for FARL versus zero
for Eureka and LiMEN at 1M steps. Figure~\ref{fig:funnel}a shows where each method stops making progress. Eureka acquires the key
in 88.0\% and opens the door in 84.8\% of episodes, but never exits; LiMEN's
selected design completes none of these stages. FARL carries progress through
all three. Under EA, completion is 1.00 for FARL and LiMEN and 0.92 for Eureka:
withholding expert design therefore separates methods that succeed when
that guidance is available.

The three candidates in Figure~\ref{fig:funnel}b show why early scores
can be misleading. All are trained for 300k steps with the same five
seed IDs; C1 and C2 use PPO, while C3 uses Advantage Actor-Critic (A2C).
The curves report exit frequency in 25k-step windows. C1 reaches 79.9\%
at 75k but falls to 5.2\% by 300k. C2 remains near 60\%, with large
differences across seeds. C3 starts at only 8.0\%, then reaches 99.5\%
at 150k and remains close to complete success. Selecting only the early
leader would miss this more reliable design.

A separate comparison keeps observations, reward, and exploration fixed
while changing the learner and its hyperparameters. Final completion
changes from 20\% to 100\% at 300k (Appendix~\ref{app:generated-example}).
The design comparison in Appendix~\ref{app:design-doorkey} shows that FARL
retains the raw input and finds a count-based reward with A2C. An interface
that fails with one training configuration can therefore work with another.

\paragraph{Ant: finding useful combinations and training them further.}
At 5M training steps per policy, FARL's EW design reaches $313.1\pm31.7$
meters, compared with Eureka's $94.1\pm50.9$ and native PPO's
$81.9\pm19.8$ (Figure~\ref{fig:funnel}d). Its forward displacement is thus
3.3 times Eureka's at this budget. The design reaches $273.3\pm15.4$
meters at 1M and $345.8\pm14.2$ at 10M. At 10M, LiMEN reaches $89.9\pm20.4$, while native
PPO falls to 37.0. More training benefits FARL's design, whereas extending
the native configuration does not recover comparable locomotion.

FARL searches Ant at budgets of 5k, 50k, 300k, and 1M steps, introducing new combinations as the search proceeds (Appendix~\ref{app:ant-designs}). Among 121 five-seed designs, the best of 42 PPO,
42 Twin Delayed Deep Deterministic Policy Gradient (TD3), and
37 Soft Actor-Critic (SAC) configurations reaches 7.4, 211.6, and 273.3 meters,
respectively. Even within SAC, results range from $-4.0$ to 273.3 meters.
Choosing SAC alone is therefore insufficient: its performance also depends on the observations, reward, and training settings. 
With expert guidance available, FARL reaches 281.1 meters at 10M, compared
with Eureka's 110.8 at 5M and LiMEN's 144.2 at 10M
(Table~\ref{tab:mainresults}).

An unexpected result is that the stronger Ant design is found without
expert guidance: 345.8 versus 281.1 meters at 10M, a 23\% increase.
The winning programs reveal different approaches. EA retains the native
27-dimensional observation and rewards forward speed using the supplied
control-cost term. EW adds 30 coordinate-difference features and rewards
forward motion together with differentiated actuator commands; both use
SAC. Appendix~\ref{app:design-ant} contrasts these programs with the native
speed--effort reward and the posture-based shaping found by the baselines. This contrast
shows that expert design need not set the limit of what automated search
can discover.

\paragraph{Hand: from occasional success to reliable relocation.}
At 5M training steps per policy, FARL achieves $\handSuccess\pm\handSD\%$
sustained relocation, compared with $1.00\pm1.00\%$ for native PPO and
zero for Eureka, LiMEN, native SAC, and native TD3
(Table~\ref{tab:mainresults}). FARL completes reach, lift, transport,
and relocation at \handReach\%, \handLift\%, \handTransport\%, and
\handSuccessRounded\%, respectively (Figure~\ref{fig:funnel}e).
LiMEN reaches the object in 88.3\% of episodes, lifts it in 16.7\%, and transports it in 1.7\%.

Earlier Hand experiments screen candidates at 5k steps and test smaller
sets at 100k and 500k. One candidate is retained despite zero success at
5k; with further training, it learns to reach and lift the object
(Appendix~\ref{app:hand-breakthrough}). In the later run that produces
the final design, iterations 0--10 use 1M steps per policy and iterations
11--15 use 5M. The search retains programs and experimental findings
between iterations, but each policy is trained from scratch.

By iteration 8, FARL has found a generated reward and 111-dimensional
representation that achieve 4.17\% success with SAC. The representation
includes palm--object and object--target distances computed from public
inputs. At iteration 11, longer training with revised SAC settings raises
success to 40.89\%, although results vary widely across seeds. Policies
can reach and lift the object, but transport remains unreliable.
Comparisons of successful and failed trajectories suggest testing whether
a simpler representation would improve this stage.

Iteration 12 tests that prediction by removing seven relational features.
The resulting 104D alternative keeps the same 84 public inputs and 20
other derived features. Both representations use the same reward,
learner, training settings, budget, and seeds. Transport reaches 38.44\%
with 104D, compared with 60.11\% with 111D: the proposed simplification
does not provide the predicted improvement. FARL records this result
and keeps both representations available for further experiments.

Iteration 13 tests new SAC settings. Final success rises to 90.89\% with
104D and \handSuccess\% with 111D, whose seed SD falls from about 51
percentage points at iteration 11 to \handSD. The selected policies
reach 90.07\% relocation frequency in the training window ending at 3M
(Figure~\ref{fig:funnel}f). Later iterations do not improve on this design,
which remains the best at the end of iteration 15. The search thus
corrects an unsuccessful representation change while continuing to
improve the training configuration (Appendix~\ref{app:hand-breakthrough}).

\paragraph{Search expenditure.}
FARL uses 82.3M training interactions on DoorKey, 1,900.7M on Ant, and
\handSteps M on Hand. Ant and Hand costs record one complete search.
DoorKey requires slightly less training than Eureka's 85M and reaches
100\% success versus zero. Ant and Hand use more search training and
produce substantially better policies. Appendix~\ref{app:search-cost}
details the expenditure.

\subsection{Core Component Ablations}\label{sec:ablation}
We remove one mechanism group at a time (Table~\ref{tab:ablation-results}).
Ablation 1 (A1) searches observations and rewards
with a fixed learner, hyperparameters, and exploration procedure. Ablation 2 (A2) retains
all design choices but uses basic feedback without rich execution evidence
or experimental memory. Ablation 3 (A3) retains joint design and diagnosis but replaces
diversity coverage, protected retention, and revival with ordinary sampling
and task-score selection. We compare these searches with previously selected FARL designs; Appendix~\ref{app:ablation-design} reports all nine cohorts and their search budgets.

\begin{table}[t]
\centering
\caption{Three core ablations in EW and frozen FARL design references.
DoorKey/Ant policy horizons are 300k/1M; task success (\%) or signed
displacement, mean $\pm$ sample SD. References use five seeds, ablations three.
Search protocols are given in Appendix~\ref{app:ablation-design}.
$^\dagger$Door reference: 50k, two seeds; ablations: 1M, three seeds.}
\label{tab:ablation-results}
{\footnotesize
\setlength{\tabcolsep}{5pt}
\begin{tabular}{@{}lccc@{}}
\toprule
Condition & DoorKey (\%) & Ant (disp.) & Door (\%)$^\dagger$ \\
\midrule
% Generated by paper/data/v12_results.py; pending is never zero.
FARL (fixed reference) & $\boldsymbol{100.00\pm0.00}$ & $\boldsymbol{273.35\pm15.35}$ & $\boldsymbol{100.00}$ \\
A1: obs.--reward search only & $0.00\pm0.00$ & $-3.09\pm5.38$ & $0.00\pm0.00$ \\
A2: no rich evidence / memory & $0.00\pm0.00$ & $-0.10\pm0.39$ & $0.00\pm0.00$ \\
A3: no diversity / preservation & $\boldsymbol{100.00\pm0.00}$ & $2.40\pm2.93$ & $0.00\pm0.00$ \\
\bottomrule

\end{tabular}}
\end{table}

\paragraph{Joint design and evidence-informed revision.}
A1 and A2 both achieve zero DoorKey completion, compared with 100\% for
FARL. Their intermediate behavior reveals different remaining failures:
A1 acquires the key in 18.0\% and opens the door in 1.0\% of episodes;
A2 reaches 46.7\% and 34.3\%, but still never exits
(Table~\ref{tab:ablation-doorkey-stages}). Allowing more design
choices gives A2 a broader search, yet basic feedback does not lead it to
a design that completes the sequence. On Ant, A1/A2 yield $-3.09$/$-0.10$
meters, versus $273.35$ for the reference. In these searches, neither
observation--reward design with a fixed solver nor joint design with basic
feedback recovers FARL's behavior.

\paragraph{Keeping alternatives available.}
A3 retains 100\% DoorKey completion but reaches only $2.40$ meters on Ant.
Its simpler selection rule finds a complete navigation strategy, while
the Ant search misses the effective designs found by the full framework.
On Door, all three restricted searches yield zero at 1M, versus 100\%
for the 50k reference. The role of each mechanism thus depends on the task:
among the restricted searches, only A3 on DoorKey matches the corresponding
FARL reference. The ablations, learning curves, and controlled Hand comparison together show how the search depends on the designs it retains and the feedback used to revise them (Figure~\ref{fig:funnel}b,f).

\FloatBarrier
\section{Discussion and Future Work}\label{sec:discussion}

The 104D Hand representation performs worse than 111D at iteration 12,
but both improve under the training settings of iteration 13
(Appendix~\ref{app:hand-breakthrough}). This is why FARL records test
conditions and keeps components available after an unsuccessful comparison.

Deciding how much training a candidate deserves remains difficult:
DoorKey includes a slow learner that becomes reliable, whereas native
Ant performance declines with longer training (Figure~\ref{fig:funnel}b,d).
Better estimates of learning potential and reuse of findings across
tasks could reduce repeated or unproductive training. Extending FARL
to visual tasks will also require diagnosis when relevant state
variables and milestones are not directly available.

\section{Conclusion}\label{sec:conclusion}
We introduced FARL, a framework that automates the joint design of RL
interfaces and solvers. FARL diagnoses learning failures, tests proposed
changes, and uses the results to decide which designs to revise or train
further. By retaining alternatives and the conditions of earlier findings,
it can revisit unsuccessful combinations without exhaustively training
the full design space. Experiments demonstrate successful design without
native expert guidance and substantial gains on locomotion and dexterous
manipulation, including \handSuccessRounded\% Hand relocation success at
5M training steps per policy.
These results extend automated RL design to the experimental decisions
needed to develop effective learning setups.

\clearpage
% Transparency and reproducibility statements.
\section*{AI Use Statement}
Generative AI assisted manuscript drafting and revision, literature search and
related-work discovery, framework and method
development, implementation and
debugging, experimental-design feedback, and interpretation of experimental
records. Figure~\ref{fig:overview} includes AI-generated illustration elements.
LLMs also form part of FARL itself, as described in Section~\ref{sec:method}.
The authors are responsible for the paper's content, results, and artifacts.

\section*{Reproducibility Statement}
Appendix~\ref{app:cohort-interfaces} specifies task interfaces;
Appendix~\ref{app:cohort-details} identifies outcome cohorts and evaluation
denominators. Appendices~\ref{app:typed-programs}--\ref{app:operational-choices} give
construction, allocation, and diagnostic rules.
Appendix~\ref{app:ablation-design} reports intervention definitions,
seed-level outcomes, and native-control settings;
Appendix~\ref{app:mechanism-evidence} links executed designs to measurements.
Appendix~\ref{app:reproducibility} describes resource accounting and the
records needed to reconstruct the results.
Source code and recorded experiment results are included in the ancillary
materials accompanying this preprint.

\clearpage
\bibliographystyle{iclr2027_conference}
\bibliography{references}

\clearpage
\appendix
\raggedbottom
\renewcommand{\thetable}{A\arabic{table}}
\renewcommand{\theHtable}{A\arabic{table}}
\setcounter{table}{0}

% Reader-oriented supplement. Source histories remain in the reproducibility
% records; only current method definitions and completed reporting cohorts enter.
\section{Task Interfaces and Evaluation}\label{app:surfaces}\label{app:formal-contract}\label{app:authority}\label{app:protocol}
The appendix describes the task interfaces and evaluation (A), design
generation (B), search and exploration (C), diagnosis and memory (D),
ablations and native controls (E), learning curves and search histories (F),
resource use and reproducibility (G), related work (H), and native and
discovered learning designs (I).

\subsection{Inputs and Access Settings}\label{app:cohort-interfaces}
The user supplies an executable task, its intent, observation/action semantics,
a task evaluator, and a resource protocol. Optional milestone predicates are
specified before search. FARL (Fully Automated Reinforcement Learning)
uses LLM-generated proposals and execution feedback to construct a learning design
$h=(\phi,r,L,\lambda,e,b)$: policy observation map, reward, learner,
hyperparameters, exploration, and learner-training depth. It does not change
the benchmark dynamics or redefine the task outcome.

\begin{table}[!t]
\centering\footnotesize
\caption{Task interfaces. EW means expert-withheld; EA means expert-available.
Hand is the stress test (ST) under EA permissions.}
\label{tab:tasks}\label{tab:regimes}
\begin{tabular}{@{}>{\raggedright\arraybackslash}p{2.1cm}>{\raggedright\arraybackslash}p{4.2cm}>{\raggedright\arraybackslash}p{5.9cm}@{}}
\toprule
Task & Benchmark identifier & Observation / action interface \\
\midrule
DoorKey & \texttt{MiniGrid-DoorKey-}\newline\texttt{5x5-v0} & EW: 148 inputs, 7 discrete actions; EA: 83 inputs, 7 actions. \\
Ant & \texttt{Ant-v4} & 27 raw inputs, 8 continuous actions; derived features are design choices. \\
Door & \texttt{door-open-v3-}\newline\texttt{goal-observable} & Meta-World; 39 inputs, 4 continuous actions. \\
Hand & \texttt{AdroitHandRelocate-v2} & ST: 84 inputs, 30 continuous actions; native control: 39 inputs. \\
\bottomrule
\end{tabular}
\end{table}

EW candidate programs consume the public transition, actions, time, and bounded
episode memory, but not native reward, privileged step metadata, or hidden
state. Task intent and field semantics remain available: withholding Ant's
native shaping does not withhold its forward-displacement objective. EA/ST
permits native feedback and the expanded task interface. In EW, the evaluator
returns aggregate outcomes during search without exposing its episode labels
to the policy or generated reward at each step. The same separation applies
to Hand's evaluation milestones. DoorKey EA additionally exposes benchmark
key-possession, door-open, and terminal-success fields; its selected reward
uses those fields (Appendix~\ref{app:design-doorkey}). The initial sensing
feedback is specified separately in Section~\ref{app:adapter-bootstrap}.

DoorKey EW flattens a $7\times7\times3$ egocentric symbolic image, divides
its categorical codes by 10, and appends direction divided by three. EA
instead exposes the entire $5\times5\times3$ grid with normalized channels,
plus eight values for agent position/direction, carried-object attributes,
elapsed time, and a constant. The smaller EA vector therefore reveals more
global information. EW--EA comparisons include this observability change,
rather than isolating native-reward access.

Hand ST appends 36 generalized velocities and the three-dimensional absolute
palm, object, and target positions to the native 39 inputs. Native controls
use only the original 39. The selected Hand policy transforms the public
84-dimensional input into 111 features; the matched comparison uses 104
(Appendix~\ref{app:hand-breakthrough}). EA baselines retain their declared wrappers: LiMEN's
DoorKey bridge exposes the full-grid vector, native reward, and time;
Eureka can receive the environment object. Hand LiMEN receives selected
numeric metadata, whereas Eureka can access the benchmark and filtered step
metadata. These are the task-specific access paths used in the EA comparisons.

All experiments use numerical observations. MiniGrid encodes object
categories, colors, and door states; locomotion and manipulation expose
physical coordinates and velocities. EW withholds the additional expert
learning artifacts, not the public observation API itself. EA/ST admits the
task-specific training feedback described above.

\subsection{Outcome and Milestone Definitions}\label{app:hand-champion-cohorts}
For a fixed design and training seed $s$, evaluation begins from the task's
original reset distribution with all source-policy assistance disabled.
With episode outcomes $\finalomr(\tau_{s,j})$, the reported seed score is
\[
 \finalomr_s(h)=\frac{1}{n_s}\sum_{j=1}^{n_s}\finalomr(\tau_{s,j}(h)).
\]
We report the equal-seed mean and sample SD, with evaluation episodes nested
within policy-training seeds. Evaluations may participate in adaptive design
selection; separately reserved fixed-design cohorts are identified below.

DoorKey and Door report task completion. Ant reports signed root displacement
until standard health termination, truncation, or 1,000 control steps.
This continuous metric measures net forward progress from the initial root
position.

Each Hand milestone must hold for five consecutive steps and is then latched
for that episode: reach requires palm--object distance $<0.10$\,m; controlled
lift requires object height $>0.04$\,m and palm--object distance $<0.12$\,m;
transport adds object--target distance $<0.15$\,m to the lift conditions;
relocate requires the native success predicate, also used for Final. These
are marginal episode events; each is scored over all evaluated episodes.
The supplied predicates specify task decomposition and thresholds.
Formally, the optional milestone set is $\stageomr=\{S_1,\ldots,S_K\}$.
For binary completion tasks, its terminal predicate equals the task outcome,
$S_K(\tau)\equiv\finalomr(\tau)$. Earlier predicates localize partial
competence without imposing an additional ordered-prefix condition on
completion. Ant's continuous displacement objective needs no milestone or
binary completion threshold. Candidate-generated progress proxies are
learning hypotheses; they do not change these supplied evaluation functions.
Figure~\ref{fig:funnel}e uses one fixed design
and one cohort per method for every stage.

\subsection{Cohorts Behind the Main Results}\label{app:cohort-details}
Table~\ref{tab:cohort-details} specifies training replication and evaluation
denominators. Horizons describe the reported policies; search costs are
additional and are accounted separately in Appendix~\ref{app:search-cost}.
Seeds replicate policy training for a selected design.

\begin{table}[!t]
\centering\footnotesize
\caption{Outcome cohorts: runs $\times$ evaluation episodes/run $\times$
training steps/run. Unless noted, runs train one selected design.}
\label{tab:cohort-details}\label{tab:calibration-cohorts}
\setlength{\tabcolsep}{4pt}
\begin{tabular}{@{}llc>{\raggedright\arraybackslash}p{5.5cm}@{}}
\toprule
Task & Method & Cohort & Selection and replication \\
\midrule
DoorKey EW & FARL & $5\times25\times1$M & Seeds 200--204; the same design has five 300k runs, seeds 100--104. \\
 & Eureka & $5\times25\times1$M & Seeds 0--4, fixed selected reward. \\
 & LiMEN & $1\times20\times1$M & Selected short-stage run; the full-stage threshold was not reached. \\
DoorKey EA & FARL & $3\times5\times100$k & Seeds 0--2; completed early-saturation cohort. \\
 & Eureka & $2\times25\times500$k & Seeds 101--102; 22/25 and 24/25 successes. \\
 & LiMEN & $3\times20\times2$M & Seeds 42--44; aggregate also used in selection. \\
Ant EW & FARL & $5\times5\times10$M & Seeds 42--46; independent 5M training uses these same IDs. \\
 & Eureka & $5\times25\times5$M & Seeds 0--4, fixed selected reward. \\
 & LiMEN & $3\times20\times10$M & Seeds 42--44; aggregate also used in selection. \\
Ant EA & FARL & $7\times3\times10$M & Seeds 0--6, one fixed configuration. \\
 & Eureka & $5\times25\times5$M & Five independent runs of the selected reward. \\
 & LiMEN & $6\times20\times10$M & Seeds 100--105; external aggregation described below. \\
Door EW & FARL & $2\times5\times50$k & One selected design. \\
 & Eureka & $3\times25\times10$M & Three successful generated designs, not seeds of one recipe. \\
 & LiMEN & $3\times20\times5$M & One selected design. \\
Hand ST & FARL & $3\times300\times5$M & Seeds 0--2; Iteration-13 selected design, also evaluated during search. \\
 & Eureka & $5\times25\times5$M & Seeds 0--4; fixed champion stage profile. \\
 & LiMEN & $3\times20\times5$M & Seeds 42--44; fixed champion stage profile. \\
\bottomrule
\end{tabular}
\end{table}

LiMEN DoorKey EW reports $0.00\pm0.00$ over its 20 zero-success evaluation
episodes from one trained candidate; its SD describes episodes. Other reported
SDs summarize training-seed means unless explicitly marked.

Door's ceiling values are reported as completion rates. Eureka's 75/75
summarizes three selected successful designs. DoorKey EA is a positive
reference at its recorded depth, using its native completion measure.
For DoorKey's fixed 1M cohort, episode $j=0,\ldots,24$ resets with seed $s+j$;
the five policy seeds therefore share some reset IDs.

\paragraph{External Ant aggregation.}\label{app:limen-ant-aggregation}
LiMEN's supplied Ant EA statistic removes one minimum and one maximum of six
seed means, then reports the central-four mean and population SD:
$144.23\pm4.65$ displacement. The same six recorded values give
$142.85\pm10.91$ under equal-seed averaging and sample SD. The main table
marks the supplied statistic explicitly. Seed IDs and 10M budgets are in
the original CSV; its evaluation script specifies 20 episodes per seed.

\section{Interface--Solver Co-Design: Construction and Execution}\label{app:typed-programs}
\subsection{Adaptation and Exploratory Learning}\label{app:adapter-bootstrap}
FARL uses an existing environment constructor or validated adapter when
available. Otherwise, it inspects candidate packages and callable signatures,
tests reset and step calls, and generates a Gymnasium-compatible adapter.
When a call fails, it uses the traceback to attempt a bounded repair.
Before generating designs, it checks observation shapes, action validity,
finite values, and short rollouts. The task must provide a usable non-visual
API. Ant EW used package discovery and adapter generation with
\texttt{gymnasium[mujoco]}; DoorKey EW, Ant EA, and Hand ST used existing
benchmark adapters.

A 500-step random probe supplies initial ranges and action responsiveness.
When enabled, a fresh miniRL policy then collects behavior: a deep Q-network (DQN) for discrete
actions and SAC for continuous actions. It trains in 10k chunks, using an
initial plan based on interface dimensions and probe statistics. After three
chunks, stabilization of observation ranges, reward summaries, or information
trigger rates can stop sensing. A failed learner falls back to random sampling.
This stopping rule tracks the stability of the collected evidence. Candidate
policies subsequently train with fresh weights and replay.

\begin{table}[!t]
\centering\footnotesize
\caption{Executed initial sensing. Follow-up steps collect generator input
with the probe policy; they are not extra optimizer updates.}
\label{tab:bootstrap-executions}
\begin{tabular}{@{}llllr@{}}
\toprule
Task & Learner & Permitted probe feedback & Learning & Follow-up \\
\midrule
DoorKey EW & DQN & Aggregate native scalar & 30k & 2k \\
Ant EW & SAC & Controllable novelty & 30k & 5k \\
Ant EA & SAC & Native scalar & 30k & 5k \\
Hand ST & SAC & Native scalar & 100k & 5k \\
\bottomrule
\end{tabular}
\end{table}

DoorKey's EW sensing phase permits aggregate native scalar feedback, while
its solving-design programs cannot consume native reward. Ant EW also
withholds that scalar from sensing and diagnosis; its probe uses novelty,
state change, and event bonuses on action-responsive public coordinates.
Thus the exact probe channel is task-specific and declared before search.
The collected transitions, value ranges, actions, and feedback guide the
first feature and reward proposals, before any successful behavior is available.

\subsection{Candidate Interfaces and Validation}
Observation proposals select semantic coordinate groups or append deterministic
expressions to the full raw vector. LLM-derived and typed-combinator expressions
can expose differences, distances, norms, ratios, or predicates. Each carries
source coordinates, type, formula, and relevance rationale. Probe checks test
numerical execution and prioritize comparisons; short-run constancy does not
justify permanently discarding a rare-event feature. These are structured
feature maps, not arbitrary stateful observation programs.
For raw observation $O_t$, the two forms are
\[
\phi_{\mathrm{sub}}(O_t)=O_t[M],\qquad
\phi_{\mathrm{aug}}(O_t)=[O_t,f_1(O_t),\ldots,f_m(O_t)],
\]
where the ordered subset $M$ is nonempty and augmentation preserves all raw
coordinates. The original raw view is a separate reference alternative.

Reward proposals provide a complete transition function and can attach a
mechanism claim, predicted behavior, a distinguishing observation, risks,
and a falsifier, as detailed below.
The runtime function reads the canonical observations before policy feature
selection, actions, time, and episode-local memory, returning scalar reward
and optional named diagnostics/memory updates. EA/ST may additionally read
native artifacts. Acquisition, persistence, regression, and event sequences
can therefore be expressed as programs, rather than coefficients in one fixed
template. Later proposal batches condition on existing families and evidence;
exact-source deduplication and bounded unproductive batches control expansion.

Learner proposals select supported implementations compatible with the action
space, together with architectures and conditional configuration domains.
Before training, the executor checks input dependencies, dimensions, action
compatibility, and valid learner parameters. Each training record identifies
the observation program, reward, learner, hyperparameters, and exploration
settings it used.

The executor requests a repair when a program misuses a declared input or
reverses its intended reward direction. It checks the relevant changes
together before training the revised design. Term weights must preserve
the declared incentive sign; named diagnostics are treated as separate
outputs unless the program explicitly includes them in its reward.
Completed seed records are retained across retries, and unfinished runs
remain marked as incomplete.

% Restored mechanism detail for Appendix B; no historical result is introduced.
\subsection{What Interface--Solver Co-Design Generates}
\label{app:v12-codesign-objects}
Interface--Solver Co-Design makes two coupled choices: what the policy sees
and is rewarded for, and how a learner acquires behavior from that interface.
Table~\ref{tab:v12-generator-objects} gives the inputs and outputs of each
generator. LLMs produce observation and reward programs and propose
compatible learners and exploration settings. The executor checks these
components, combines them into a complete design, and trains a policy with
the selected RL algorithm. Recording the components separately allows
later experiments to identify exactly which choice changed.

\begin{table}[htbp]
\centering\footnotesize
\setlength{\tabcolsep}{4pt}
\renewcommand{\arraystretch}{1.10}
\caption{Executable objects in joint design. Inputs follow the task's
declared access setting; available execution evidence informs later proposals.}
\label{tab:v12-generator-objects}
\begin{tabular}{@{}>{\raggedright\arraybackslash}p{.16\textwidth}
>{\raggedright\arraybackslash}p{.35\textwidth}
>{\raggedright\arraybackslash}p{.42\textwidth}@{}}
\toprule
Design axis & Evidence supplied & Generated output \\
\midrule
Observation & Coordinate meanings/types, task relations, probe transitions,
and earlier feature comparisons & Ordered field selections or raw inputs
augmented by deterministic expressions, with source dependencies and formulas. \\
Reward & Task intent, transition interface, observed behavior, current
incentives, and retrieved explanations & Complete transition-reward source
and a testable account of its proposed behavioral mechanism. \\
Learner & Action-space type, representation, supported implementations,
and learning-health evidence & Compatible algorithm and network alternatives
with learner-conditioned configuration domains. \\
Training & Selected interface/learner, exploration alternatives, completed
learning evidence, and resource protocol & Instantiated hyperparameters,
exploration settings, and requested training depth for each execution. \\
\bottomrule
\end{tabular}
\end{table}

Observation construction retains raw inputs as an explicit alternative.
The generator groups coordinates by meaning and proposes both small subsets
and broader selections. LLM proposals and combinations of typed expressions
add features such as displacement, distance, and threshold events. Each
expression is tested on collected samples before it is added to a candidate.
These checks help prioritize tests, while retaining features that may matter
only during rare events. Reward programs continue to receive the original
transition observations regardless of the policy's feature map.

\paragraph{A reward program carries a prediction.}
Reward programs can encourage reaching a useful state, remaining there,
avoiding a reversal, or completing a sequence of events. Episode-local memory
allows rewards to depend on earlier events. A proposal can also state why
its program should help and what behavior would test that explanation
(Table~\ref{tab:v12-reward-prediction}). Explanatory fields are optional;
the generated source defines the reward actually used in training.

\begin{table}[htbp]
\centering\footnotesize
\setlength{\tabcolsep}{4pt}
\renewcommand{\arraystretch}{1.10}
\caption{The prediction record attached to a reward proposal. Source code
defines the executed incentive; the other fields guide subsequent tests.}
\label{tab:v12-reward-prediction}
\begin{tabular}{@{}>{\raggedright\arraybackslash}p{.28\textwidth}
>{\raggedright\arraybackslash}p{.67\textwidth}@{}}
\toprule
Record & Question it makes explicit \\
\midrule
Program and family & What incentive is executed, and which proposed
mechanism distinguishes this alternative? \\
Mechanism and behavior & Why might it help, and which events or behavior
should change if it works? \\
Distinguishing prediction & Which observation separates this explanation
from a competing explanation? \\
Falsifier and proxy risks & Which result would challenge the mechanism,
and how might reward improve without the intended behavior? \\
Temporal role & Does the proposal emphasize acquisition, maintenance,
regression prevention, or their interaction? \\
Compatibility and depth & Which learner properties and minimum training
duration are proposed for an informative test? \\
\bottomrule
\end{tabular}
\end{table}

For example, consider a hand that sometimes lifts an object but never
completes relocation. It may need to lift more reliably or to hold the object
longer after lifting. A reward for initiating a lift predicts more lifts;
a reward for maintaining the grasp predicts longer holds. Measuring both
would distinguish these explanations even before either policy completes
the task. This illustrative test specifies the behavior to compare when
the final score alone gives little guidance.

\section{Diversity-Preserving Search and Frontier Exploration}\label{app:algorithm}
\subsection{Controller and Allocation Rules}\label{app:retention-tests}
\begin{algorithm}[htbp]
\caption{FARL's experimental loop}\label{alg:aspire}
\footnotesize
\begin{algorithmic}[1]
\REQUIRE task/interface, fixed outcome evaluator, optional milestones, budget ladder with seed allocations, and stopping protocol
\STATE adapt the environment; collect permitted exploratory evidence
\STATE initialize executable component pools, candidate bank, experimental memory, and cost ledger
\WHILE{the stopping protocol permits another experiment}
\STATE retrieve relevant findings, unresolved tests, and candidate learning evidence
\STATE propose and validate components; resolve required construction/learner repairs
\STATE reserve diagnostic comparisons and controls; fill remaining capacity by component-family and pair coverage
\STATE assign new designs a validity or learning rung; select retained designs for deeper trials or replication using outcomes, learning dynamics, and experimental need
\STATE bind any frontier sources, reset mixture, withdrawal schedule, and additional interaction cost
\STATE fix exact designs, conditional hyperparameters, seeds, and training budgets for each comparison
\STATE train independent learners; evaluate task outcomes and milestones; record effective settings and costs
\STATE check executed changes and conditions against the requested experiments
\STATE update conditional findings; retain competitive, improving, uncertain, or experimentally necessary designs and reusable components
\STATE use these findings to request the next comparison, component revision, or budget promotion
\ENDWHILE
\STATE select using the declared task-outcome statistic; execute any reserved fixed-design cohort
\RETURN selected design, seed outcomes, experimental records, stopping reason, and accounted cost
\end{algorithmic}
\end{algorithm}

The scheduler uses a successive-halving-style budget ladder, with the
retention rules below preserving learning potential and required comparisons.
A valid initial probe is not a sufficient basis for rejecting a design's
learning quality. Greater depth can be assigned to a new training run of a
retained design; promotion does not require continuation of its earlier
policy parameters. The ladder and stopping protocol are declared separately:
training depth alone does not determine whether an entire search has ended.

Coverage uses keys for learner identity, observation and exploration types,
reward programs or primitives, and component families. Required controls
enter first. Greedy selection then adds candidates covering the largest
number of previously uncovered components and pairs, breaking ties
deterministically. Remaining slots go to underrepresented families.
The records distinguish proposed combinations from those actually trained.

At a validity rung, valid low-scoring designs receive no learning-quality
pruning. Later bounded-task retention considers alternative reasons: any
positive final score ($>10^{-12}$), positive milestone evidence
($>10^{-6}$) with rising/inconclusive/high-variance dynamics, improving training,
unavailable curves, or a required control/protected incumbent. For at least four
finite returns, let $a,b$ be first-/last-quarter means and $R_{\min},R_{\max}$
the sequence extrema. Training improvement is
\[
 g_s=(b-a)/\max(1,|a|,|b|,R_{\max}-R_{\min}),\qquad
 \operatorname{median}_s g_s>10^{-3}.
\]
Unavailable data are not negative evidence. Returns are compared within a
program, not used to rank different reward programs.

Milestone dynamics use early/late windows of width
$q=\min\{\max(10,\lfloor N/4\rfloor),\max(1,\lfloor N/2\rfloor)\}$.
Fewer than 40 episodes is inconclusive. Otherwise, for rates $p_a,p_b$,
$\Delta=p_b-p_a$ has diagnostic band
$\Delta\pm1.96\sqrt{[p_a(1-p_a)+p_b(1-p_b)]/q}$, clipped to $[-1,1]$.
A lower endpoint above 0.05 is rising; an upper endpoint below $-0.05$ is
collapse; $|\Delta|\leq0.05$ with zero in the band is plateau; other cases
are inconclusive. Across seeds, a rate range of at least 0.35 takes priority
as high variance; otherwise a strict majority determines the state. These
normal-approximation bands diagnose learning behavior, not an exact sequential
test. A requested target stage takes precedence over the first available stage.
Plateau/collapse can end unchanged milestone-based continuation at a completed
learning rung, while other retention reasons remain applicable.

For unbounded outcomes with $n\geq3$ scores, scheduling uses
$\ell_i=\mu_i-0.5s_i/\sqrt n$ and $u_i=\mu_i+0.5s_i/\sqrt n$.
The largest-$\ell$ candidate, candidates with $u_i\geq\max_j\ell_j$, and
required controls remain. With fewer than three scores, the mean is retained
but variance is unreliable, giving temporary protection. These bands are
allocation heuristics, not calibrated confidence intervals; screening may
pool multiple configurations and repeated seed IDs. Bounded reward--outcome
pathology checks additionally route completed high-return/zero-outcome designs
without stage or learning improvement to revision. Their return-sign test is
scale-dependent, so the behavioral evidence is essential to interpretation.

The hyperparameter optimization (HPO) incumbent uses the arithmetic mean for fewer than four scores.
Otherwise, it averages sorted zero-based indices $\lfloor n/4\rfloor$ through
$\lfloor3n/4\rfloor$ inclusive. Round ties retain the first eligible design;
later rounds replace it only on strict improvement. Final reporting still
uses the equal-seed mean and sample SD. Explicit comparison protocols may
specify another selection statistic; DoorKey's three-design promotion used
a normal-approximation lower bound, then its recorded trimmed statistic and
mean. Archived components can replenish depleted pools without transferring
policy parameters, optimizer state, or replay to new training units.

\subsection{Predicate-Guided Frontier Exploration}\label{app:frontier-exploration}
A policy that reaches part of the task can help another learner explore
the next stage. The controller first verifies its saved checkpoint and
compatible action and observation interfaces. Occasional success is enough
to try it as a source policy; routine reuse requires the specified evidence
across seeds and checkpoints.

Before handing control to the learner, the controller checks the current
state and remaining episode time against the source's readiness condition.
Each source policy is assessed separately. If its condition cannot be
evaluated from available quantities, the handoff does not proceed.

One frozen source acts from an ordinary reset through legal environment steps.
On current readiness, control passes to the fresh trial learner; source actions
do not enter the learner's rollout or replay and no parameters are fused.
For $u$ learner steps completed of $b$, the default assistance probability is
$p_{\mathrm{assist}}(u)=0.5\max\{0,1-u/(0.8b)\}$. The final training tail
therefore uses only original resets. At most three source attempts are made
per reset, and a handoff must leave at least one quarter of the original
horizon. The horizon is not restarted or extended. Sources and schedule are
fixed within a worker; completed waves can inform subsequent candidates.

If the source fails to reach the required state, training falls back to a
fresh reset and the source remains available. Invalid actions, corrupt
checkpoints, and incompatible inputs require repair. All source interactions,
successful or not, count in addition to learner steps. Final evaluation
disables assistance and evaluates one learner from original resets, with
freshly initialized stage detectors. The reported champions train from
ordinary resets.

% Component preservation and frontier decisions for Appendix C.
\subsection{Preserving Options Beyond One Unsuccessful Pairing}
\label{app:v12-preservation-options}
Diversity-Preserving Search retains both complete designs and reusable
components. A poor result applies to the tested combination at its training
budget and initialization; the reward may still work with another learner.
Diagnosis can therefore propose a different pairing or further training.
The scheduler uses these requests and the observed learning progress to
allocate the available budget.

The scheduler first reserves required controls and protected designs.
It then favors candidates containing component families or pairs that have
not yet been covered, and fills any remaining capacity from less represented
families. Thus two candidates sharing a reward can receive training with
different learners. Their measured behavior determines which combinations
merit further search.

The bank stores each component's source, parent, status, and test results.
Retained components can be recombined; when an active component pool falls
below its configured minimum, archived bank entries can replenish it. The
default minimum population and maximum revival count are both two.
This replenishment makes earlier components available for new combinations.
Ablation 3 disables it. Recombined designs retain their component history
and train fresh policies.

\subsection{Using Intermediate Behaviors to Support Exploration}
\label{app:v12-frontier-decisions}
A policy that occasionally reaches a milestone can help another learner
explore what comes next. It need not yet be reliable enough for repeated
reuse. Table~\ref{tab:v12-frontier-decisions} separates these uses of partial
behaviors. Final evaluation always starts from the task's original resets
with assistance disabled.

\begin{table}[htbp]
\centering\footnotesize
\setlength{\tabcolsep}{4pt}
\renewcommand{\arraystretch}{1.10}
\caption{Different evidence warrants different uses of partial competence.}
\label{tab:v12-frontier-decisions}
\begin{tabular}{@{}>{\raggedright\arraybackslash}p{.29\textwidth}
>{\raggedright\arraybackslash}p{.29\textwidth}
>{\raggedright\arraybackslash}p{.35\textwidth}@{}}
\toprule
Available evidence & Next use & Question for the next experiment \\
\midrule
No compatible source with usable readiness & Joint design or exploration
from original resets & Observe partial competence and identify a testable
frontier opportunity. \\
Positive local signal; checkpoint and runtime interface verifiable &
Exploratory source-policy attempts mixed with original resets & Whether the
source can reliably create that opportunity and the new learner can use it. \\
The same source also satisfies its declared repeated-use checks &
Qualified reuse under the recorded checkpoint/interface and scope & Whether
the complete learner succeeds without assistance. \\
Completed whole-task evidence for one learner & Deeper training of the
selected recipe or reserved fixed-design evaluation & Performance at the
newly evaluated depth and replication level. \\
\bottomrule
\end{tabular}
\end{table}

\paragraph{Choosing source policies.}
Some candidates always train from original resets. Others receive
assistance from compatible source policies, with priority given to sources
that more reliably reach the target milestone. Rotation among source
families, every four allocations by default, also gives weaker policies
further opportunities. Controlled comparisons use a fixed source plan
across the compared designs.

Appendix~\ref{app:frontier-exploration} specifies when control passes from
the source policy to the learner, how assistance decreases during training,
and how the additional environment interactions are counted.

\section{Falsifiable Diagnosis and Memory}\label{app:operational-choices}\label{app:router}
\subsection{From Observation to an Executable Test}
Diagnosis considers the training return, task score, milestones, trajectories,
and optimizer activity together. A failed evaluation first needs repair.
A completed evaluation with zero success raises a different question:
whether the policy needs more exploration, a different representation, or
longer training. Table~\ref{tab:router} gives examples of such decisions.

\begin{table}[!t]
\centering\footnotesize
\caption{Examples of observations and the comparisons they motivate.}
\label{tab:router}
\begin{tabular}{@{}>{\raggedright\arraybackslash}p{3.5cm}>{\raggedright\arraybackslash}p{4.2cm}>{\raggedright\arraybackslash}p{4.5cm}@{}}
\toprule
Observed discrepancy & Next comparison & Discriminating observation \\
\midrule
Frontier event never observed & Exploration change with design basis fixed & Increased event coverage or frontier entry. \\
Useful events observed but poorly represented & Public feature/observation comparison & Better separation of relevant states and subsequent behavior. \\
Reward improves without task progression & Alternative incentive or learner test & Predicted behavioral transition, not merely larger return. \\
Behavior improves but varies across seeds & Replication or greater depth & Consistent progress under the stated conditions. \\
Training progress is absent from saved evaluation & Same-checkpoint, same-episode evaluator check; then retention test & Separate measurement disagreement from irreproducible behavior. \\
Declared learner updates do not occur & Learner/exploration repair & Changed effective settings and measured optimizer activity. \\
\bottomrule
\end{tabular}
\end{table}

A hypothesis names supporting observations, scope, alternatives, a predicted
effect, a falsifier, and the component to change. The trajectory planner,
internal diagnosis, and critic submit proposals to one controller; disagreement
can request a distinguishing experiment. A controlled comparison binds an
unchanged reference and a changed-axis design, common seed sets and depth,
and compatible non-target components. The executor can use a verified paired-seed
schedule or repeated matched trial groups. Within-learner comparisons share
their declared base hyperparameters; an exploration treatment transforms its
compatible settings. Learner changes use typed, algorithm-specific domains,
not an identical-hyperparameter claim across different algorithms.

Before interpreting a comparison, the controller checks the executed
components, seeds, interface, evaluation status, and training steps.
A missing comparison or a revision that makes no change leaves the
hypothesis unresolved. Construction repairs do not require training a
known-invalid control. When the learner is not updating correctly, the
controller checks its settings and optimizer steps, repairs the fault, and
retains any valid behavioral observations. Repaired candidates can re-enter
search after satisfying the same checks as other designs.

\subsection{Retrieval, Updating, and Test Priority}
Experimental memory stores explanations, predictions, scope, evidence, and
unresolved comparisons; the candidate bank separately stores trainable
components. Retrieval deduplicates hypothesis IDs in stored traversal order,
selects the same access setting, and orders states as supported, suspected,
proposed, inconclusive, falsified, and exhausted, with recency breaking ties.
The controller lesson view admits up to 12 entries and the generator view up
to 16 within its context envelope. Current designs and recent trial summaries
occupy separate context sections. Retrieving a falsified explanation preserves
its scope rather than banning every use of its underlying program family.

Observational support can make a hypothesis suspected. Stronger ledger states
require the appropriate claim type, linked comparison cohorts, valid information
use, and repeated directional agreement. A complete paired-seed execution
record admits actual outcomes from at least two distinct paired seeds. Without
that record, the bound-trial producer averages seeds within matched trial
groups and requires at least two groups. A higher treatment mean and a strict
majority of positive differences support the local prediction; reverse evidence
contradicts it. Reusing a seed across groups does not create new independent
seeds. Conflicting or missing comparisons remain inconclusive. These are operational evidential states,
not calibrated posterior probabilities. Public-event association with a
milestone and the effect of rewarding that event are separate hypotheses.

The trajectory planner ranks admissible tests lexicographically by summed
discriminating-value score (default one per addressed explanation), number
of explanations addressed, and lower ordinal cost; exact ties keep library
order. A concrete public-goal-interface alternative receives a priority bonus
of 100 over a generic perturbation of the same goal. A minimum-learning-depth
guard prevents treating a validity probe as a sufficient performance test.

When no authoritative request is available, a compact-memory fallback uses
$q(h)=g(h)f(h)[0.5+u(h)]v(h)/c(h)$: recorded discriminating value $g$ and
positive cost $c$ default to one; $f$ is one with a falsifier and 0.25 without;
$u=\max(0.1,1-\mathrm{confidence})$. State weights are 1.2 (supported),
0.9 (suspected), 0.75 (proposed/live), 0.65 (inconclusive), and 0.6 (weakened).
Testing, dormant, falsified, and exhausted entries are not actionable in this
fallback. These planning scores are heuristics. Neither route bypasses
execution checks or revives a consumed request through an older memory view.

\subsection{Conditional Training Configuration}
Optuna's tree-structured Parzen estimator (TPE) uses sampler seed $42+t$ at controller iteration $t$. The learner-
conditioned domain can be proposed from task evidence. Fallback on-policy
domains are log learning rate $[10^{-4},3\times10^{-3}]$, discount
$[0.95,0.999]$, log entropy coefficient $[10^{-4},0.05]$, and generalized advantage estimation (GAE) parameter
$[0.9,0.99]$; off-policy fallbacks use log learning rate $[10^{-4},10^{-3}]$,
the same discount range, log target-update rate $[0.005,0.02]$, and batch size
$\{128,256,512\}$. Each value must belong to the selected learner's executable
domain; a cached proposal for another learner is not a valid configuration.

Local contraction requires an explicit optimization request and healthy
completed evidence from the same executable recipe and training conditions.
It halves support in the parameter's linear or logarithmic sampling coordinate
around the compatible best trial, clipped to the original domain; discrete
domains remain unchanged. A flat nonpositive outcome field preserves support,
as do new designs, unresolved frontier exploration, health repairs, and missing
compatible evidence. Iteration number is not a shrink factor. Exploration's
learner-specific effects and executable warmup rules are detailed in
Appendix~\ref{app:v12-exploration-effects}.

% Mechanism detail for Appendix D; terminology follows V13.
\subsection{Connecting Diagnosis, Experiments, and Memory}
\label{app:v12-experiment-binding}
Falsifiable Diagnosis tests an explanation before recording it as a finding.
The candidate bank stores reusable programs, configurations, and their
history. Scientific memory stores explanations together with their
predictions, results, and test conditions. The LLM uses behavioral summaries
and earlier findings to propose a comparison; the controller selects its
components, runs it, and checks the outcome.
Table~\ref{tab:v12-test-binding} follows these steps from the proposal to
the memory update.

\begin{table}[htbp]
\centering\footnotesize
\setlength{\tabcolsep}{4pt}
\renewcommand{\arraystretch}{1.10}
\caption{From a proposed diagnostic test to a result that can guide later experiments.}
\label{tab:v12-test-binding}
\begin{tabular}{@{}>{\raggedright\arraybackslash}p{.18\textwidth}
>{\raggedright\arraybackslash}p{.39\textwidth}
>{\raggedright\arraybackslash}p{.36\textwidth}@{}}
\toprule
Object & Required content & Role in the next decision \\
\midrule
Proposed test & Explanation, competing account, changed axis, prediction,
falsifier, and requested depth & Select a comparison that can discriminate
between explanations. \\
Bound comparison & Exact reference and treatment artifacts, compatible
non-target stack, seed IDs, and declared replication unit & Reserve executable
arms and bind learner-appropriate hyperparameter settings. \\
Completed evidence & Materialized settings, actual updates/depth, evaluation
validity, and outcomes attached to the bound arms & Determine whether the
requested contrast actually occurred and has interpretable results. \\
Memory update & Local prediction verdict, its scope, and links to the
comparison and observations & Inform subsequent retrieval, revision, or
another discriminating test. \\
\bottomrule
\end{tabular}
\end{table}

A controlled comparison changes one design choice while holding the other
compatible components fixed. Exploration variants can modify a shared set
of base hyperparameters. Comparisons between different learners instead use
each algorithm's valid hyperparameter domain. The recorded constructor
settings identify the configuration actually used in training.

A verified paired-seed schedule can use one matched trial group with at least
two distinct paired seeds. It requires completed arms, design and interface
identities that match the declared comparison, and actual per-seed outcomes; a trial average cannot be
expanded into artificial seed observations. Without this complete record, the
bound-trial producer uses at least two matched trial groups and their seed
means. The recorded replication unit governs interpretation; reusing a seed
across groups does not create another independent seed. Optional frontier
initialization is bound with the comparison rather than rotated between arms.
Repairs of proved construction errors follow a separate path: a known-invalid
control need not consume a learning budget.

Observed behavior can mark an explanation as \emph{suspected} and prompt a
test. Supporting a prediction requires the completed comparison to agree
both in its average effect and in a strict majority of paired seeds or
matched trial groups. An effect in the opposite direction contradicts the
prediction. Missing comparisons, a revision that changes nothing, or
conflicting outcomes leave the question unresolved. These memory states
summarize the experiments; they are not posterior probabilities.

Retrieval keeps the test conditions with each finding. A reward that fails
with one learner can still be tried with another. Similarly, accepting a
repair or allocating more training is a search decision; the resulting
experiment determines whether it supports the explanation.

\subsection{Exploration Settings}
\label{app:v12-exploration-effects}
Interface--Solver Co-Design chooses exploration settings alongside the
learner. Table~\ref{tab:v12-exploration-effects} lists the supported options.
Execution records contain the resulting training configuration.

\begin{table}[htbp]
\centering\footnotesize
\setlength{\tabcolsep}{4pt}
\renewcommand{\arraystretch}{1.10}
\caption{Supported exploration settings for each learner.
Here $b$ is the requested number of learner-training steps.}
\label{tab:v12-exploration-effects}
\begin{tabular}{@{}>{\raggedright\arraybackslash}p{.24\textwidth}
>{\raggedright\arraybackslash}p{.71\textwidth}@{}}
\toprule
Choice & Executed effect and eligibility \\
\midrule
Native exploration & No additional modification to algorithm-native behavior. \\
Entropy profile & PPO entropy coefficient at least 0.005;
SAC selects adaptive-entropy initialization 0.1 or 0.2. \\
Coverage profile & PPO entropy coefficient at least 0.01;
SAC adaptive-entropy initialization 0.2 or 0.5; compatible off-policy
warmup proposal $\max(1000,0.25b)$. \\
Long warmup & SAC/TD3 or deep deterministic policy gradient (DDPG) warmup proposal $\max(5000,0.20b)$,
subject to the executable learning-start guard. \\
Controllable novelty & Training-only bonus over measured action-responsive
observation coordinates; a targeted variant uses available candidate-goal
coordinates. \\
Public-goal replay descriptor & Requires public goal atoms; the recorded
implementation does not activate SB3 hindsight experience replay (HER) merely by selecting this descriptor. \\
\bottomrule
\end{tabular}
\end{table}

SAC values initialize an adaptive coefficient rather than impose an entropy
floor. A profile with no implementation branch for the selected learner
has no extra effect; for example, the coverage profile does not impose an
A2C entropy floor. Valid explicit warmup values below $b$ are honored.
Missing values or settings leaving no update opportunity invoke the
short-rung fallback; invalid values require correction. Intrinsic bonuses
modify training incentives; evaluation always uses the original task metric.

\section{Component Ablations and Native Controls}\label{app:ablation-design}
\subsection{Three Core Interventions}
\paragraph{Ablation 1: restrict search to observations and rewards.}
Observation/reward generation, diagnosis, and candidate management remain.
The learner is fixed to PPO with two 64-unit layers, default hyperparameters,
native exploration, and a predefined training ladder. Its configuration is
fixed independently of the optimized FARL champion. This comparison tests
the benefit of searching over learner and training choices alongside the
observation and reward.

\paragraph{Ablation 2: remove rich evidence and experimental memory.}
All design axes and diversity management remain. Diagnosis, planning, arbitration,
and revision receive the task, public interface, current design, execution
status, and basic training/Final feedback. Detailed trajectories, milestone
changes, coverage analyses, reward--outcome consistency, causal-comparison
records, and retrieved experimental lessons are excluded, including derived
summaries. Initial public-interface perception and independent audit logging
remain available. This tests rich evidence and memory as one mechanism group.

\paragraph{Ablation 3: remove diversity and conditional preservation.}
All design axes and evidence-informed diagnosis remain. Novelty-oriented
generation, family/pair coverage, protected lineages, and historical revival
are removed. Candidates are sampled from current component pools without
diversity quotas. All valid candidates remain eligible at the validity rung;
at learning rungs, the ordinary bank retains the top $\lceil n/3\rceil$ of
$n$ eligible candidates by its task-outcome commitment score. Required
experimental controls remain. A family may recur by chance, but discarded
candidates cannot be deliberately revived through history.

\subsection{Search and Evaluation Protocol}
Each condition searches with its specified mechanism group enabled or
removed, then fixes one selected design for independent policy training.
DoorKey and Ant use seeds 200--202 at 300k and 1M steps per seed,
respectively, with five final evaluation episodes per seed. Door uses the
same three seeds at 1M steps and 25 evaluation episodes per seed.
Tables~\ref{tab:ablation-search-allocation} and \ref{tab:ablation-seeds}
report search allocations and all final seed outcomes.

\begin{table}[!t]
\centering\footnotesize
\caption{Recorded search training for the ablation comparisons, in millions
of interactions; final fixed-design training is accounted separately.
Each column uses independently run searches with the indicated allocations.}
\label{tab:ablation-search-allocation}
\begin{tabular}{@{}lrrr@{}}
\toprule
Condition & DoorKey & Ant & Door \\
\midrule
FARL reference & 72.800 & 1,825.735 & 9.100 \\
A1 & 21.960 & 56.860 & 64.055 \\
A2 & 17.760 & 8.460 & 3.060 \\
A3 & 5.560 & 51.960 & 10.060 \\
\bottomrule
\end{tabular}
\end{table}

\begin{table}[!t]
\centering\footnotesize
\caption{All final seed-level outcomes underlying Table~\ref{tab:ablation-results}.
DoorKey and Door entries are percent; Ant entries are signed displacement.}
\label{tab:ablation-seeds}
\begin{tabular}{@{}llrrr@{}}
\toprule
Task & Intervention & Seed 200 & Seed 201 & Seed 202 \\
\midrule
% Generated by paper/data/v12_results.py. DoorKey/Door: percent; Ant: signed displacement.
DoorKey & A1 & $0.00$ & $0.00$ & $0.00$ \\
 & A2 & $0.00$ & $0.00$ & $0.00$ \\
 & A3 & $100.00$ & $100.00$ & $100.00$ \\
\addlinespace[2pt]
Ant & A1 & $-6.7831$ & $3.0899$ & $-5.5665$ \\
 & A2 & $-0.3452$ & $0.3557$ & $-0.3055$ \\
 & A3 & $5.7242$ & $1.2957$ & $0.1823$ \\
\addlinespace[2pt]
Door & A1 & $0.00$ & $0.00$ & $0.00$ \\
 & A2 & $0.00$ & $0.00$ & $0.00$ \\
 & A3 & $0.00$ & $0.00$ & $0.00$ \\
\bottomrule

\end{tabular}
\end{table}

The FARL references use the selected designs in
Appendix~\ref{app:cohort-details}. DoorKey and Ant align the policy-training
horizon across conditions; references use five training seeds and ablations
three. DoorKey's reference uses 25 evaluation episodes per seed and its
ablations five. Door's reference uses two 50k policies; its ablations use
three 1M policies. Ant A2 retains its existing reward program at the
revision step. These comparisons evaluate the designs found by each
mechanism configuration under the reported search allocations.

\paragraph{Where zero completion arises.}
The final DoorKey policies also have a separate 100-episode stage evaluation
per seed (Table~\ref{tab:ablation-doorkey-stages}). These are marginal rates
over all episodes, not conditional on completing earlier stages. A1 seldom opens the door; A2 more often acquires the key and opens the door
but never reaches the exit. Their zero final scores therefore arise from
different remaining transitions. A3 completes the sequence in 98.67\% of
this stage-evaluation sample and 100\% of the five-episode final evaluation.

\begin{table}[!t]
\centering\footnotesize
\caption{DoorKey ablation stage diagnostics, percent, mean $\pm$ seed SD.
All three seeds train for 300k; each stage evaluation uses 100 episodes.}
\label{tab:ablation-doorkey-stages}
\begin{tabular}{@{}lrrr@{}}
\toprule
Intervention & Pick up key & Open door & Reach exit \\
\midrule
% Generated from final-policy stage evaluations, 100 episodes/seed.
A1 & $18.00\pm17.09$ & $1.00\pm1.73$ & $0.00\pm0.00$ \\
A2 & $46.67\pm7.23$ & $34.33\pm17.67$ & $0.00\pm0.00$ \\
A3 & $98.67\pm2.31$ & $98.67\pm2.31$ & $98.67\pm2.31$ \\
\bottomrule

\end{tabular}
\end{table}

Door uses the 5k/50k/300k/1M search ladder. Final training keeps the selected
observation map, reward, learner, and hyperparameters fixed. All three
interventions complete their final evaluations, with zero successful
episodes in each seed. The separate DoorKey stage evaluation describes
behavior after search and is not used to select the final design.

\subsection{Hand Native-Interface Long Training}\label{app:native-control}
This control uses native 39-dimensional observations and canonical
dense reward with fixed SB3 2.8.0 configurations (Table~\ref{tab:native-config}), without search, normalization,
added features, demonstrations, or source-policy assistance. The environment is
\texttt{AdroitHandRelocate-v2}, 30 actions, 200-step horizon; its implementation
is pinned to upstream commit \texttt{62640af45fc1} (full identity in the manifest).
Results are given in Table~\ref{tab:native-results}.
PPO and SAC use library defaults; TD3 additionally uses the official example's
independent Gaussian action noise with standard deviation 0.1. These are
standard library configurations, not an Adroit-specific tuned preset.

\begin{table}[!t]
\centering\footnotesize
\caption{Fixed native-control learner settings. Unlisted settings retain
the version-pinned library defaults; full constructor receipts are supplied.}
\label{tab:native-config}
\begin{tabular}{@{}lccc@{}}
\toprule
Setting & PPO & SAC & TD3 \\
\midrule
Hidden layers & $64,64$ (Tanh) & $256,256$ (ReLU) & $400,300$ (ReLU) \\
Learning rate / discount & $0.0003/0.99$ & $0.0003/0.99$ & $0.001/0.99$ \\
Batch size & 64 & 256 & 256 \\
Rollout / epochs & 2048 / 10 & --- & --- \\
Replay / warmup & --- & $10^6/100$ & $10^6/100$ \\
Train frequency / gradient steps & --- & $1/1$ & $1/1$ \\
Entropy coefficient & 0 & automatic & --- \\
Target-update rate & --- & 0.005 & 0.005 \\
Action noise & policy sampling & policy sampling & Gaussian, $\sigma=0.1$ \\
\bottomrule
\end{tabular}
\end{table}

Each learner has three independently initialized trajectories (seeds 0--2),
trained continuously through nominal 1M and 5M checkpoints. These are two
measurements of the same trajectories, not six independent seeds. Each
checkpoint is evaluated deterministically on 100 episodes per seed, reset
with $500000+1000s+j$, $j=0,\ldots,99$. PPO checkpoints follow completed
rollout updates at 1,001,472 and 5,001,216 actual steps; SAC/TD3 use exact
1M/5M. The nine trajectories consume 45,003,648 training interactions.

\begin{table}[!t]
\centering\footnotesize
\caption{Completed native Hand controls; the PPO 5M stage profile is plotted in
Figure~\ref{fig:funnel}e. Stage and Final entries are percent,
equal-seed mean $\pm$ sample SD. Final requires five consecutive native-success
steps; Any is at least one successful step in the episode.}
\label{tab:native-results}
\setlength{\tabcolsep}{3pt}
\begin{tabular}{@{}llrrrrr@{}}
\toprule
Learner & Steps & Reach & Lift & Transport & Final & Any \\
\midrule
PPO & 1M & $70.67\pm9.45$ & $26.00\pm6.93$ & $1.67\pm1.53$ & $0.67\pm1.15$ & $1.67\pm2.89$ \\
 & 5M & $42.00\pm3.61$ & $9.33\pm4.04$ & $1.00\pm1.00$ & $1.00\pm1.00$ & $3.00\pm1.00$ \\
SAC & 1M & $25.33\pm8.02$ & $4.00\pm2.00$ & $0.00\pm0.00$ & $0.00\pm0.00$ & $0.00\pm0.00$ \\
 & 5M & $18.00\pm9.54$ & $5.33\pm4.16$ & $0.67\pm1.15$ & $0.00\pm0.00$ & $0.67\pm1.15$ \\
TD3 & 1M & $8.00\pm10.39$ & $0.67\pm1.15$ & $0.00\pm0.00$ & $0.00\pm0.00$ & $0.33\pm0.58$ \\
 & 5M & $0.00\pm0.00$ & $0.00\pm0.00$ & $0.00\pm0.00$ & $0.00\pm0.00$ & $0.00\pm0.00$ \\
\bottomrule
\end{tabular}
\end{table}

The corresponding native returns at 1M/5M are $-1.60\pm3.84$/$-10.49\pm5.29$
(PPO), $-14.69\pm4.03$/$-11.42\pm0.72$ (SAC), and
$-9.95\pm0.48$/$-7.82\pm1.45$ (TD3). PPO's Final successes at 1M are
0/100, 0/100, and 2/100; at 5M they are 0/100, 2/100, and 1/100.
SAC and TD3 have zero Final successes in every seed at both checkpoints.
PPO's lift rate falls from 26.00\% to 9.33\% while transport remains near 1\%:
additional optimization does not steadily accumulate the required behaviors.
TD3's improving native return likewise does not imply relocation success.

These paired training-depth controls use the native 39-dimensional interface
and a fixed configuration. FARL constructs a 111-dimensional policy input
from its 84-dimensional public interface and performs design search. At 5M,
its selected design reaches $\handSuccess\pm\handSD\%$ sustained relocation,
compared with native PPO's $1.00\pm1.00\%$. The stage profiles compare the
resulting behaviors (Figure~\ref{fig:funnel}e).

\subsection{Ant Native-Reward Training}\label{app:native-ant}
The Ant native control uses PPO with the environment's native dense reward
and 27-dimensional observations. Three independent policy seeds (0--2)
follow one training trajectory each, with evaluation checkpoints at 1M, 5M,
and 10M environment steps. Each checkpoint is evaluated deterministically
on 25 episodes per seed using the same signed root-displacement metric as
Figure~\ref{fig:funnel}d: final minus initial root position, to termination
or the 1,000-step cap. Reset seeds are $10000+100s+j$, $j=0,\ldots,24$.

Table~\ref{tab:native-ant-results} reports the outcomes.
The fixed PPO configuration uses eight environments, 256 steps per environment
per rollout, batch size 64, ten optimization epochs, learning rate $3\times10^{-4}$,
$\gamma=0.99$, GAE $\lambda=0.95$, clipping 0.2, entropy coefficient 0,
and separate $64\times64$ Tanh policy and value networks. There is no
observation/reward normalization, interface search, or policy assistance.
Table~\ref{tab:native-ant-results} reports completed checkpoint evaluations;
all model hashes, constructor settings, and episode outcomes are retained.

\begin{table}[!t]
\centering\footnotesize
\caption{Native-reward PPO on Ant. Per-seed means use 25 evaluation episodes;
the aggregate is the equal-seed mean $\pm$ sample SD.}
\label{tab:native-ant-results}
\begin{tabular}{@{}lrrrr@{}}
\toprule
Steps/seed & Seed 0 & Seed 1 & Seed 2 & Displacement \\
\midrule
1M & \textbf{113.08} & 37.57 & 36.41 & $62.36\pm43.93$ \\
5M & 96.86 & \textbf{59.41} & \textbf{89.40} & $\boldsymbol{81.89\pm19.82}$ \\
10M & 21.45 & 50.34 & 39.17 & $36.99\pm14.57$ \\
\bottomrule

\end{tabular}
\end{table}

\section{Executed Search and Design Cases}\label{app:mechanism-evidence}\label{app:detailed-results}
This section describes the designs found in search, how their policies
learn, and how the Hand representation experiment changes the controller's
next decision.

\subsection{Ant: Design Diversity and Performance}\label{app:ant-designs}
The executed search inventory counts distinct component artifacts at each
training depth using all 1,554 completed valid worker results from the frozen EW archive.
It excludes proposed-but-untrained combinations. At 5k, 50k, 300k, and 1M
steps per seed, respectively, the numbers of completed design evaluations
are 27, 56, 1,350, and 121. The corresponding distinct counts are
$(6,6,5,5)$ observations, $(18,18,18,18)$ rewards, $(3,3,3,3)$ learners,
and $(6,6,5,4)$ exploration artifacts. An artifact may occur at several
depths; these counts should not be summed into a unique-design total.
These counts describe the alternatives that receive training at each budget.

Figure~\ref{fig:funnel}c shows \emph{every} completed five-seed 1M design in the final
search cohort: 42 PPO, 42 TD3, and 37 SAC configurations. A dot is a design's
equal-seed mean displacement.
Horizontal jitter only separates overlapping dots. Diamonds mark the highest
mean within each learner, 7.40, 211.56, and 273.35. Observation, reward,
exploration, and hyperparameters also vary within learner groups; the plot
therefore exposes the range of joint-design outcomes. All 121 designs use
1M per seed and five evaluation episodes per seed, including designs with
zero or negative displacement.

\begin{table}[!t]
\centering\footnotesize
\caption{Selected Ant designs and their complete EW/EA configurations.}
\label{tab:ant-executed-designs}
\begin{tabular}{@{}>{\raggedright\arraybackslash}p{2.9cm}>{\raggedright\arraybackslash}p{4.4cm}>{\raggedright\arraybackslash}p{4.7cm}@{}}
\toprule
Component & EW & EA \\
\midrule
Policy input & 27 raw + 30 difference features & 27 raw coordinates \\
Reward & Forward-motion credit, action-diversity terms, coordinate-norm penalties & Speed/control-cost bands plus continuous velocity and cost terms \\
Learner / network & SAC / $256,256$ & SAC / $256,256,128$ \\
Optimizer learning rate & $2.5981\times10^{-4}$ & $3\times10^{-4}$ \\
Entropy / warmup & Automatic, initial 0.2 / 10k & Automatic, default / 20k \\
Replay & Ordinary replay & Ordinary replay \\
\bottomrule
\end{tabular}
\end{table}

Table~\ref{tab:ant-budget-detail} lists the budget-specific outcomes.
The EA design retains the native 27-dimensional observation and reuses
forward-velocity and control-cost metadata. With $v$ denoting
\texttt{x\_velocity} and $c=-\texttt{reward\_ctrl}$, its reward is
\[
 r_{\mathrm{EA}}=0.8b(v,c)+0.2v-0.15c,\qquad
 b(v,c)=\begin{cases}
 1 & v>1.0,\ c<0.9,\\
 0.5 & v>0.6,\ c<0.7\ \text{otherwise},\\
 0 & \text{otherwise}.
 \end{cases}
\]
Thus, EA builds on native reward components without copying the native total
reward. EW instead adds 30 coordinate-difference features and computes reward
from the public observation and actions. Its terms credit forward motion,
variance across actuator commands, and differences between two groups of
actuator commands, with penalties on selected coordinate norms. These
programs express different objectives: EA rewards speed within a control-cost
budget, whereas EW directly encourages differentiated actuation alongside
forward motion. Both selected designs use SAC and ordinary replay, with the
network and training settings in Table~\ref{tab:ant-executed-designs}.

The search histories also differ. EA iterations 1--5 retain this same
observation--reward pair while testing learners and hyperparameters. EW
continues to evaluate five observation artifacts and 18 reward artifacts
at 1M. Its stronger final result therefore comes from a design that departs
from the supplied speed--effort reward structure.

\begin{table}[!t]
\centering\footnotesize
\caption{Ant outcomes by training budget in Figure~\ref{fig:funnel}d.
Rows summarize recorded training cohorts; EA 1M includes six search configurations.}
\label{tab:ant-budget-detail}
\begin{tabular}{@{}llccr@{}}
\toprule
Setting & Steps/seed & Seeds & Episodes/seed & Displacement \\
\midrule
EW & 1M & 0--4 & 5 & $273.3\pm15.4$ \\
EW & 5M & 42--46 & 5 & $313.1\pm31.7$ \\
EW & 10M & 42--46 & 5 & $345.8\pm14.2$ \\
EA & 1M & $6\times(0\text{--}4)$ & 5 & $207.1\pm85.6$ \\
EA & 10M & 0--6 & 3 & $281.1\pm74.4$ \\
\bottomrule
\end{tabular}
\end{table}

EW rows train the selected design independently at each budget. The five
10M-minus-5M differences average 32.72. The EA 1M point aggregates 30
configuration--seed results from six hyperparameter settings of the selected
structural family; EA 10M uses the final selected configuration. Lines in
Figure~\ref{fig:funnel}d summarize outcomes by budget.

\subsection{DoorKey: One Reward, Different Training Configurations}\label{app:generated-example}
The frozen selected program reads the 49 object-category slots in the public
148-dimensional input. For category code $c$, its smooth visible count is
$C_c(O)=\sum_{i\in\{0,3,\ldots,144\}}\max(0,\min(1,1-10|O_i-c|))$.
The codes 0.5/0.4/0.8 denote key/door/goal. Pickup can earn 2.0 when the visible
key count drops by more than 0.6 from above 0.8; otherwise it receives 0.05
times the count decrease. Toggle earns 0.15 times the absolute door-count
change. Forward movement with a goal-count increase above 0.6 earns 0.8;
otherwise the goal term is 0.04 times its count increase. Each step also
incurs $-0.0015$. This is a visibility-based shaping program, not an oracle
for key possession, door opening, or exit completion. The exact generated
source (SHA-256 prefix \texttt{07251b7d0ca4}) is included in the artifact package.

\begin{table}[!t]
\centering\footnotesize
\caption{Same observation, reward, and exploration artifacts; different
training configurations. Both use five seeds (100--104), 300k steps and
25 evaluation episodes per seed.}
\label{tab:doorkey-config}
\begin{tabular}{@{}lrr@{}}
\toprule
Setting & Selected A2C & PPO comparison \\
\midrule
Learning rate & $1.8485447659\times10^{-4}$ & $1.3505226478\times10^{-4}$ \\
Discount & 0.9957992252 & 0.9974352405 \\
Entropy coefficient & 0.00022588495 & 0.02108114065 \\
GAE parameter & 0.9032534459 & 0.9120453326 \\
Success by seed & $(1,1,1,1,1)$ & $(0,1,0,0,0)$ \\
\bottomrule
\end{tabular}
\end{table}

Both algorithm artifacts declare $128,128$ ReLU actor/critic networks. The
learner and the four listed hyperparameters change together, so this evidence
concerns the training configuration, not the update rule alone. In particular,
the shared coverage-profile artifact does not impose an A2C entropy floor.

\subsection{Policy Learning and Candidate Selection}\label{app:learning-curves}
\begin{figure}[htbp]
\centering
\includegraphics[width=\textwidth]{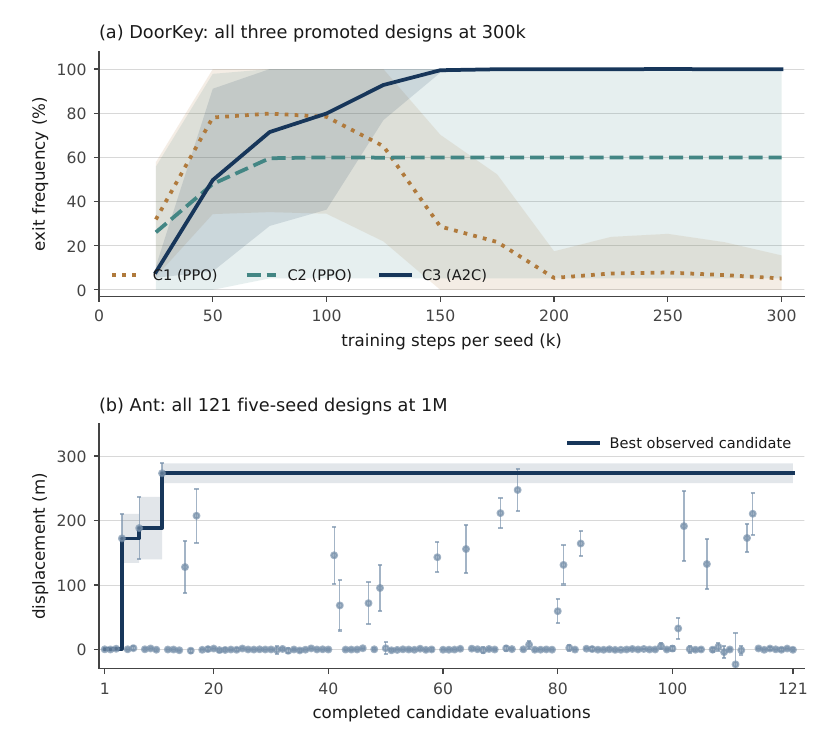}
\caption{Candidate learning and selection at fixed training budgets.
(a) DoorKey: five-seed mean exit frequency and sample SD for each of three
300k candidates. (b) Ant: all 121 candidate evaluations at 1M, with their
five-seed SDs; the step line tracks the incumbent. The horizontal axes are
policy-training steps in (a) and completed candidate evaluations in (b).}
\label{fig:candidate-learning}
\end{figure}

Figure~\ref{fig:funnel}b shows how DoorKey candidates learn as
policy-training steps increase. Figure~\ref{fig:candidate-learning} expands
this view and records Ant candidate comparisons at a fixed training budget.
The Hand learning trajectories are detailed in
Appendix~\ref{app:hand-breakthrough}.

\paragraph{DoorKey: three candidates at the same training depth.}
Curve 1 (C1) tracks design \texttt{17728f964a0e203d},
Curve 2 (C2) design \texttt{449c5b344a476e98}, and Curve 3 (C3) design \texttt{9a831d11e7f30408}.
All use five training seeds (100--104) and 300k steps per seed. C1/C2 use
PPO and C3 uses A2C. The accompanying records contain each candidate's
complete configuration. C1--C3 identify candidates, not search iterations.

DoorKey ends an episode when the agent reaches the goal or after 250 steps.
The episode logs therefore identify goal-reaching episodes shorter than
this limit. We assign completed
episodes to nonoverlapping 25k-step windows, count episodes shorter than
250 steps, and average within each seed. Curves show the equal-seed mean;
shading shows sample SD across the five seed means, clipped to the possible
frequency range $[0,1]$. An exit exactly at step 250 is conservatively counted
as timeout because the monitor does not distinguish it from truncation.
The first plotted measurement covers the first 25k training steps.
Table~\ref{tab:mainresults} reports a separate evaluation of C3 after 1M
steps, using five independently trained policies.

\paragraph{Ant: candidate comparison within the 1M allocation.}
The controller uses iteration indices 0, 1, 2, and 3 for the
5k, 50k, 300k, and 1M allocations, respectively. All 121 five-seed 1M
designs belong to iteration 3. Each dot in Figure~\ref{fig:candidate-learning}b is the
fixed-evaluator mean for one completed design, ordered by worker completion
time; parallel workers can finish in a different order from their launch.
The step line shows the largest candidate mean observed so far, and its
band is that candidate's sample SD over five policy-training seeds.

The eventual champion, \texttt{e24e4c2fa358fdcf}, is the eleventh completed
evaluation and reaches $273.35\pm15.35$ meters. It remains the incumbent
through all 121 evaluations. All candidate outcomes, including zero and
negative displacement, are plotted. Ant displacement is measured after each
policy finishes training, so the horizontal axis counts completed candidate
evaluations. Figure~\ref{fig:funnel}d separately compares the outcomes at
1M, 5M, and 10M steps per policy.

\begin{figure}[htbp]
\centering
\includegraphics[width=\textwidth]{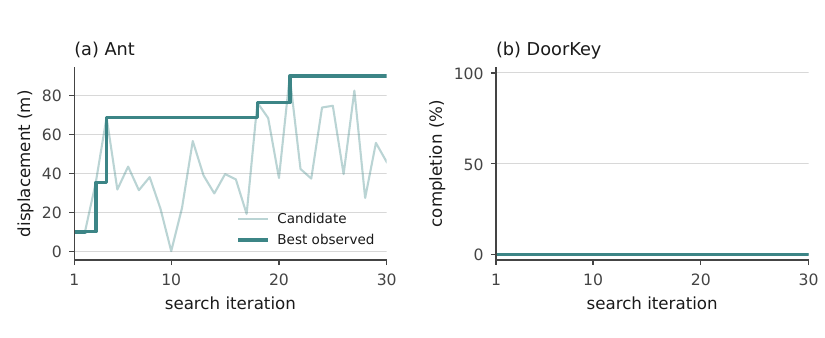}
\caption{All 30 LiMEN search evaluations in Ant and DoorKey. Faint lines
connect candidate results; dark lines show the best observed result.
The horizontal axis counts search iterations.}
\label{fig:limen-search}
\end{figure}

\paragraph{Additional baseline search records.}
Figure~\ref{fig:limen-search} gives all 30 LiMEN search evaluations for Ant
and DoorKey, together with the best result found by each iteration. Ant
evaluates each candidate after 10M steps per seed and DoorKey after 1M.
The Ant incumbent reaches the 89.9-meter
result reported in Table~\ref{tab:mainresults}.

\subsection{Hand: Reliable Relocation Through Successive Experiments}\label{app:hand-breakthrough}
\paragraph{Earlier Hand search at shorter budgets.}
Hand investigations include candidate training below 1M steps. In the
August 29 search archive, the controller first evaluates a broad population
at 5k steps, then runs smaller sets of experiments at 100k and 500k
(Table~\ref{tab:hand-early-budgets}). These are executed candidate-training
budgets, distinct from the environment-sensing phase. The counts below come
from completed worker results and their per-seed interaction receipts,
rather than from the configured maximum training depth.

\begin{table}[htbp]
\centering\footnotesize
\caption{Completed short-budget Hand experiments in the August 29 archive.
A design evaluation is one candidate configuration trained on its assigned
seeds; repeated evaluations of a component combination are counted
separately. The 500k row contains 26 completed evaluations across three
rounds. Training steps exclude evaluation interactions.}
\label{tab:hand-early-budgets}
\setlength{\tabcolsep}{4pt}
\begin{tabular}{@{}lrrrp{.34\textwidth}@{}}
\toprule
Steps/seed & Design evaluations & Seeds/design & Seed runs & Search role \\
\midrule
5k & 256 & 1 & 256 & Broad executable-design probe \\
100k & 12 & 2 & 24 & Short-budget learning comparisons \\
500k & 26 & 3 & 78 & Deeper tests of retained designs \\
\bottomrule
\end{tabular}
\end{table}

One retained design, \texttt{031b6a159bb610d7}, appears at all three
depths and subsequently at 1M. After its zero-success 5k evaluation,
the scheduler records that a valid probe should be retained without judging
its eventual learning quality. Its completed 100k evaluations also have
zero final success, while later 500k evaluations record occasional
relocation. Thus the archive documents an actual decision to give an
initially unsuccessful design further training opportunities. The respective
5k, 100k, and 500k evaluations use 3, 30, and 100 episodes per seed;
their scores are not joined into a single policy-learning curve.

This archive belongs to an earlier search with different generated
component recipes. The September 13 run that produces the reported
97.11\% champion starts without importing earlier candidates, frontiers,
experimental memory, or policy-training state. Within that later run,
the designs and evidence found at 1M are retained for the 5M experiments
described below, while the selected policies are initialized independently.
The early archive therefore establishes the use of short-budget Hand search;
the 1M-to-5M case establishes the development of the reported champion.

\begin{figure}[htbp]
\centering
\includegraphics[width=\textwidth]{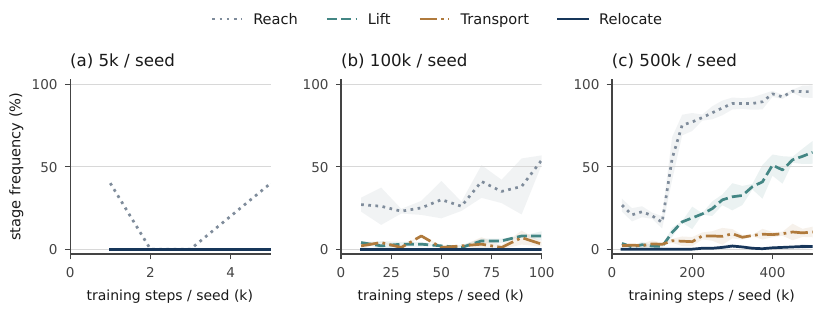}
\caption{Early Hand learning within three independently trained cohorts of
the retained design \texttt{031b6a159bb610d7}: iteration/trial 0/165 at
5k, 1/1 at 100k, and 4/1 at 500k. Curves are marginal stage-completion
frequencies from all recorded training episodes, using nonoverlapping
1k-, 10k-, and 25k-step windows, respectively. Each seed contributes
25, 500, or 2,500 episodes in the corresponding panel. Lines average
equally across the one, two, or three training seeds; bands show sample
SD where multiple seeds are available. Each panel starts with a fresh
policy and uses its own executed training configuration. The panels
are separate training histories, rather than checkpoints of one
continuously trained policy; saved-policy evaluations are reported
separately.}
\label{fig:hand-early-training}
\end{figure}

Figure~\ref{fig:hand-early-training} shows partial competence emerging before
reliable task completion. In the final 25k-step window of the 500k cohort,
reach and lift occur in 95.5\% and 58.7\% of training episodes, whereas
transport and relocation remain at 10.4\% and 1.6\%. These distinct stages
give subsequent diagnosis more information than the low final success
frequency alone.

\paragraph{From early search to the selected design.}
The run that produces the reported design uses 1M training steps per policy
at iterations 0--10, followed by 5M steps at iterations 11--15. At
iteration 8, the 111-dimensional representation, generated reward, and SAC
learner achieve 4.17\% mean success across two seeds. Later iterations
reuse these programs and the findings from earlier experiments, while
each policy starts with new weights and an empty replay buffer.

Each of iterations 11--15 tests two configurations on seeds 0--2, with
300 final evaluation episodes per seed. All ten configurations and thirty
policy-training runs are complete. The prescribed search ends after
iteration 15 without a further 10M stage. Table~\ref{tab:hand-search-cohorts}
reports every configuration. The best is trial 0 of iteration 13, genome
\texttt{1b3455f44977bd4d}; its three policies each train for 5M steps from
ordinary resets. Table~\ref{tab:hand-outcome} compares their performance
and the search cost with the automated baselines and native learners.

\begin{table}[!t]
\centering\footnotesize
\caption{Hand outcomes and expenditure. A successful seed has at least one
successful evaluation episode. Success is the equal-seed mean $\pm$ SD;
policy depth is per seed. FARL costs cover the complete search; $^*$ denotes estimated tokens.}
\label{tab:hand-outcome}
\setlength{\tabcolsep}{4pt}
\begin{tabular*}{\textwidth}{@{\extracolsep{\fill}}lrrrr@{}}
\toprule
Method & Policy depth & Successful seeds & Success (\%) & Tokens / steps (M) \\
\midrule
FARL & 5M & $\boldsymbol{3/3}$ & $\boldsymbol{\handSuccess\pm\handSD}$ & $\handTokens^*\,/\,\handSteps$ \\
Eureka & 5M & $0/5$ & $0.00\pm0.00$ & $0.284\,/\,110.4$ \\
LiMEN & 5M & $0/3$ & $0.00\pm0.00$ & $0.607\,/\,55.0$ \\
\midrule
Native PPO & 5M & $2/3$ & $1.00\pm1.00$ & $\mathbf{0}\,/\,\mathbf{15.0}$ \\
Native SAC & 5M & $0/3$ & $0.00\pm0.00$ & $\mathbf{0}\,/\,\mathbf{15.0}$ \\
Native TD3 & 5M & $0/3$ & $0.00\pm0.00$ & $\mathbf{0}\,/\,\mathbf{15.0}$ \\
\bottomrule
\end{tabular*}
\end{table}

\begin{table}[!t]
\centering\footnotesize
\caption{All ten Hand configurations evaluated at iterations 11--15 with 5M training steps per policy.
Each row uses three training seeds and 300 final evaluation episodes per
seed. Values are sustained relocation percentages; SD is across training
seeds. The 104D and 111D designs use SAC; the 84D designs use PPO except
iteration~15 / trial~1, which uses SAC.}
\label{tab:hand-search-cohorts}
\begin{tabular}{@{}lrrrrr@{}}
\toprule
Iteration/trial & Input dim. & Seed 0 & Seed 1 & Seed 2 & Mean $\pm$ SD \\
\midrule
11/0 & 104 & 4.00 & \textbf{98.67} & 0.33 & $34.33\pm55.74$ \\
11/1 & 111 & 98.00 & 0.00 & 24.67 & $40.89\pm50.97$ \\
12/0 & 111 & 0.67 & 97.33 & 74.67 & $57.56\pm50.55$ \\
12/1 & 104 & 1.33 & 92.33 & 12.33 & $35.33\pm49.67$ \\
13/0 & 111 & \textbf{99.00} & \textbf{98.67} & \textbf{93.67} & $\boldsymbol{97.11\pm2.99}$ \\
13/1 & 104 & 87.33 & 94.00 & 91.33 & $90.89\pm3.36$ \\
14/0 & 111 & 0.00 & 0.00 & 0.00 & $0.00\pm0.00$ \\
14/1 & 84 & 0.00 & 0.00 & 0.00 & $0.00\pm0.00$ \\
15/0 & 84 & 0.00 & 0.00 & 0.00 & $0.00\pm0.00$ \\
15/1 & 84 & 0.00 & 0.33 & 0.00 & $0.11\pm0.19$ \\
\bottomrule

\end{tabular}
\end{table}

The selected policies complete 297, 296, and 281 of their 300 evaluation
episodes, respectively: $\handSuccess\pm\handSD\%$ across seeds.
Evaluation starts from the original reset distribution and uses one saved
policy per seed. The final milestone uses the native sustained-success
criterion. These evaluations also guide candidate selection
(Appendix~\ref{app:hand-champion-cohorts}); Table~\ref{tab:hand-selected-stages}
reports the corresponding milestone completion rates.

\begin{table}[!t]
\centering\footnotesize
\caption{The selected Hand cohort: marginal completion of each milestone
over all 300 evaluation episodes per seed. These are not conditional
transition probabilities.}
\label{tab:hand-selected-stages}
\begin{tabular}{@{}lrrrr@{}}
\toprule
Evidence & Reach & Lift & Transport & Final \\
\midrule
Seed 0 & 100.00 & 100.00 & 100.00 & 99.00 \\
Seed 1 & 100.00 & 99.67 & 99.00 & 98.67 \\
Seed 2 & 100.00 & 97.33 & 94.33 & 93.67 \\
Mean $\pm$ SD & $100.00\pm0.00$ & $99.00\pm1.45$ & $97.78\pm3.02$ & $97.11\pm2.99$ \\
\bottomrule
\end{tabular}
\end{table}

\paragraph{Learning configurations.}
The 111D representation contains 84 public input coordinates, twenty
generated distance features, and seven additional relational features.
The 104D alternative removes the last seven while keeping the others.
Both use the same generated transition-reward program.

The seven features include palm--object, object--target, and palm--target
distances, their sum, difference, and ratio, and a coordinate difference.
For the public input vector $x$, the three distances are
$\|(x_{30},x_{31},x_{32})\|_2$, $\|(x_{36},x_{37},x_{38})\|_2$, and
$\|(x_{33},x_{34},x_{35})\|_2$, respectively, using zero-based indices.
They make approach and target relations explicit using existing inputs.
The comparison therefore changes how information is represented without
adding new sensory information. Table~\ref{tab:hand-executed-config}
lists the SAC settings at iterations 11--13.

\begin{table}[!t]
\centering\footnotesize
\caption{SAC settings at iterations 11--13. Within each iteration,
the two representations share the listed SAC settings.}
\label{tab:hand-executed-config}
\begin{tabular}{@{}lccc@{}}
\toprule
Setting & Iteration 11 & Iteration 12 & Iteration 13 \\
\midrule
Network & $384,384,256$ & $384,384,256$ & $384,384,256$ \\
Learning rate & $1.21296\times10^{-4}$ & $1.96651\times10^{-4}$ & $1.23911\times10^{-4}$ \\
Discount & 0.987555 & 0.982773 & 0.987258 \\
Target update $\tau$ & 0.003684 & 0.005543 & 0.009802 \\
Warmup steps & 200k & 1M & 1M \\
Batch size / replay capacity & $256/10^6$ & $256/10^6$ & $256/10^6$ \\
Entropy coefficient & Automatic & Automatic & Automatic \\
Learner steps / seed & 5M & 5M & 5M \\
\bottomrule
\end{tabular}
\end{table}

The representation, reward program, and network remain fixed while the
controller tests different SAC hyperparameters. Final success increases from
$40.89\pm50.97\%$ to $57.56\pm50.55\%$ to
$\handSuccess\pm\handSD\%$. At iterations~12 and~13, both representations use the same training settings
and ordinary-reset procedure, yielding direct comparisons.

\paragraph{Testing a proposed representation change.}
After iteration 11, policies reliably reach and lift the object, but
transport remains the first unresolved stage. The diagnostic record
compares fourteen trajectories with higher task outcomes and two with
lower outcomes. Their minimum values of $\mathrm{obs}[85]$ differ, with
the same direction observed in two seeds. Transport completion in this
diagnostic cohort is 43.67\%. This small observational comparison suggests
testing the representation; it does not establish the cause of the failure.

The proposed change removes seven relational features, predicting that
the simpler representation will improve transport. Iteration 12 compares
the 104D alternative with the 111D reference using the same reward,
learner, hyperparameters, exploration, budget, and seed IDs. Transport
completion is 38.44\% for 104D and 60.11\% for 111D; final success is
35.33\% and 57.56\%, respectively. The transport difference is $-21.67$
percentage points, with two of three paired seeds showing a decline.
The change therefore fails to produce the predicted improvement under
these settings. This comparison tests the seven features as a group;
it neither establishes that each is necessary nor refutes the original
observational association. Table~\ref{tab:hand-evidence-chain} summarizes
the comparison and the resulting memory update.

\begin{table}[!t]
\centering\footnotesize
\caption{The Hand representation experiment, from its initial prediction
to the measured outcome and the resulting memory update.}
\label{tab:hand-evidence-chain}
\begin{tabular}{@{}>{\raggedright\arraybackslash}p{.21\textwidth}
>{\raggedright\arraybackslash}p{.72\textwidth}@{}}
\toprule
Record & Observed content \\
\midrule
Diagnosis & Transport is the earliest unresolved stage; an observation-feature contrast motivates a representation test. \\
Experiment & Iteration 12 tests two representations with identical reward, learner, training settings, exploration, 5M depth, and seeds 0--2. \\
Prediction & The proposed representation improves transport completion. \\
Measurement & Reference/alternative transport: 60.11/38.44\%; final success: 57.56/35.33\%. \\
Conclusion & The result contradicts the predicted improvement; the reference remains champion. \\
Memory & The associated hypothesis changes from testing to falsified, citing the matched experimental contradiction. \\
\bottomrule
\end{tabular}
\end{table}

The controller marks the prediction as falsified in memory and retains
both representations for further tests. At iteration 13, new SAC settings
raise success to \handSuccess\% with 111D and $90.89\pm3.36\%$ with 104D.
The earlier comparison favors 111D under its tested settings, while the
later result shows that the lower-scoring representation can also learn
with a different configuration.

\paragraph{Later search and the retained design.}
Iteration~14 tests a new training configuration for the 111-dimensional
SAC design and a raw-input PPO design; both complete with zero final
success. Iteration~15's raw-input PPO and SAC candidates reach zero and
$0.11\pm0.19\%$ final success, respectively.
The iteration~13 policies remain the selected cohort after iteration 15 at
$\handSuccess\pm\handSD\%$. These later candidates are independent
training runs; the saved selected policies are retained throughout the search.
Table~\ref{tab:hand-later-config} lists their training settings.
The completed search through iteration~15 accounts for 408.45M environment
interactions, including 26.78M prefix interactions.
The 135 recorded LLM calls use approximately 2.58M tokens. Table~\ref{tab:hand-outcome}
reports expenditure through selection of the iteration~13 design.

\begin{table}[!t]
\centering\scriptsize
\caption{Executed settings for the later completed Hand candidates.
Every policy trains for 5M learner steps. The generated reward program is
shared; prefix interactions are summed over three seeds. A dash marks a
setting that does not apply to the learner. Column headings give iteration/trial.}
\label{tab:hand-later-config}
\setlength{\tabcolsep}{3pt}
\begin{tabular*}{\textwidth}{@{\extracolsep{\fill}}lcccc@{}}
\toprule
Setting & 14/0 & 14/1 & 15/0 & 15/1 \\
\midrule
Learner / input dimension & SAC / 111 & PPO / 84 & PPO / 84 & SAC / 84 \\
Network & $384,384,256$ & $512,512,256$ & $512,512,256$ & $384,384,256$ \\
Learning rate & $4.87439\times10^{-4}$ & $1.95026\times10^{-4}$ & $1.10072\times10^{-4}$ & $1.17979\times10^{-4}$ \\
Discount & 0.990106 & 0.967368 & 0.994906 & 0.994028 \\
Batch size & 256 & 64 & 128 & 128 \\
Entropy coefficient & Automatic & 0.032534 & 0.005 & Automatic \\
SAC target update $\tau$ & 0.005334 & --- & --- & 0.008957 \\
SAC warmup / replay capacity & $10^6/10^6$ & --- & --- & $10^4/10^6$ \\
PPO rollout length / epochs & --- & $2048/10$ & $4096/5$ & --- \\
PPO clipping / trace decay $\lambda$ & --- & $0.2/0.922306$ & $0.191816/0.962284$ & --- \\
Prefix interactions & 245,787 & 0 & 0 & 0 \\
\bottomrule
\end{tabular*}
\end{table}

\paragraph{Learning curves.}
The curves shown here cover twenty-seven 5M policy-training runs from nine
configurations. We order
the training episodes by their recorded sequence and assign completed episodes
to nonoverlapping 100k-step windows. For each seed and window, the metric
is the fraction of episodes satisfying the sustained-relocation criterion.
Lines show the equal-seed mean; bands show sample SD, clipped to
$[0,100]\%$. The first point covers the first 100k training steps.
Table~\ref{tab:hand-search-cohorts} separately evaluates the saved policies
after training.

\begin{figure}[htbp]
\centering
\includegraphics[width=\textwidth]{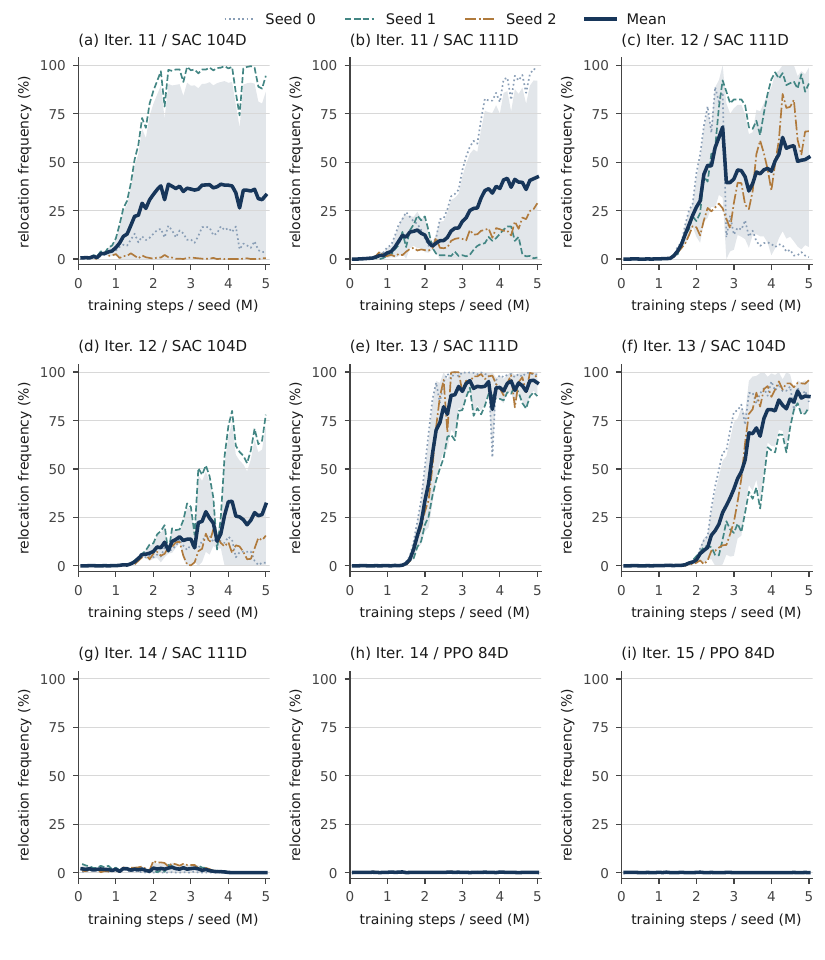}
\caption{Learning curves for nine of the ten completed 5M Hand configurations.
The final raw-input SAC configuration is reported in Table~\ref{tab:hand-search-cohorts}.
Thin curves are the
three training seeds, the thick curve is their mean, and shading is sample
SD. Each point aggregates a 100k-step training window. Iteration~12 panels
(c--d) form the matched representation experiment. Panel (e) is the selected
design; panels (g--i) show the later completed alternatives.}
\label{fig:hand-learning}
\end{figure}

Figure~\ref{fig:funnel}f follows the 111-dimensional SAC design across
iterations 11--14, including the unsuccessful configuration at iteration 14.
Figure~\ref{fig:hand-learning} shows these nine configurations and their seeds;
Table~\ref{tab:hand-search-cohorts} additionally reports the final raw-input
SAC configuration and covers all configurations through iteration 15.
The first three 111-dimensional cohorts train from ordinary resets. iteration~11's
104-dimensional candidate additionally uses 1,251,980 source-policy prefix
interactions across its three seeds; search expenditure includes these
interactions, while the learning curves count learner steps. The iteration~14
SAC candidate uses another 245,787 prefix interactions; all final evaluations
use the original reset distribution.

\begin{figure}[htbp]
\centering
\includegraphics[width=\textwidth]{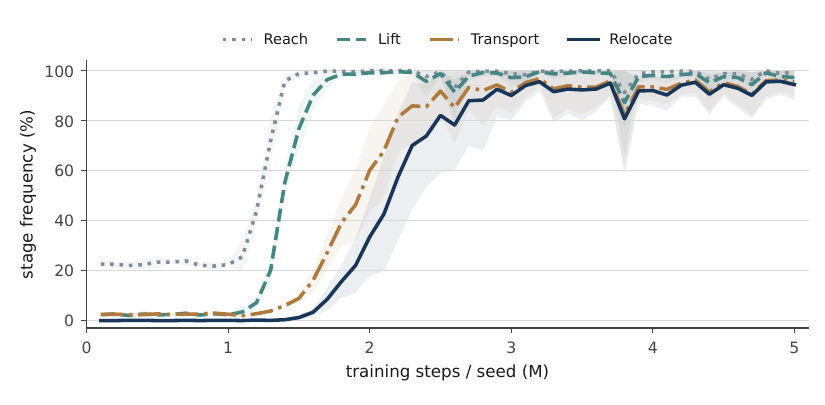}
\caption{Learning progress of the Hand design selected at iteration~13. Lines
and bands are three-seed means and SDs of 100k-step training frequencies.
The objective completion curve shows how training extends beyond lifting
and transport to sustained relocation.}
\label{fig:hand-stage-learning}
\end{figure}

The design selected at iteration~13 reaches 90.07\% mean relocation frequency by the
window ending at 3M steps and 94.47\% in the final 100k window. The last
window's seed frequencies are 98.8\%, 87.6\%, and 97.0\%.
The separate saved-policy evaluation reaches \handSuccessRounded\%
(Table~\ref{tab:hand-training-windows}).
Figure~\ref{fig:hand-stage-learning} shows learning extending from reach
and lift to transport and sustained relocation.

\begin{table}[!t]
\centering\footnotesize
\caption{Selected Hand design during training. Each row summarizes the
100k-step window ending at the indicated budget, as a percentage averaged
equally over three seeds. Final evaluation uses 300 episodes per seed.}
\label{tab:hand-training-windows}
\begin{tabular}{@{}lrrrr@{}}
\toprule
Window ending & Reach & Lift & Transport & Final \\
\midrule
1M & 22.40 & 2.27 & 2.53 & 0.07 \\
2M & 100.00 & 99.20 & 60.13 & 33.47 \\
3M & 98.80 & 97.20 & 91.40 & 90.07 \\
4M & 99.53 & 98.13 & 93.53 & 92.07 \\
5M & 99.40 & 97.33 & 95.53 & 94.47 \\
Final evaluation & \handReach & \handLift & \handTransport & \handSuccessRounded \\
\bottomrule
\end{tabular}
\end{table}

\paragraph{Earlier independent experiment.}
An earlier 1M experiment retrieved 14 lessons and generated a proximity-gated
palm-to-target reward. Its three policies achieved 5/300, 10/300, and 3/300
final successes ($2.00\pm1.20\%$). This result comes from a separate,
earlier run. Its generated program and experiment records are retained
with the project history.

\section{Reproducibility and Resource Accounting}\label{app:reproducibility}\label{app:search-cost}\label{app:claims}
\subsection{Search Costs and Language-Model Configuration}
Table~\ref{tab:mainresults} combines task outcomes with EW and Hand ST expenditure.
Table~\ref{tab:search-cost} gives the corresponding accounting. Training counts
use distinct FARL training receipts, baseline completed-unit budgets, or actual
recorded steps where available; evaluation rollouts are separate.

\begin{table}[!t]
\centering\footnotesize
\caption{EW and Hand ST resource accounting. Training steps are in millions. Tokens are
recorded input/output usage, except the rounded Ant FARL total estimate.
Hand FARL's main-table estimate additionally accounts for one missing request.}
\label{tab:search-cost}\label{tab:llm-usage}
\setlength{\tabcolsep}{4pt}
\begin{tabular}{@{}llrrrr@{}}
\toprule
Task & Method & Search steps & Extra training & Input tokens & Output tokens \\
\midrule
DoorKey & FARL & 72.800 & 9.500 & 2,629,000 & 205,025 \\
 & Eureka & 80.000 & 5.000 & 78,116 & 170,675 \\
 & LiMEN & 30.000 & 0 & 983,594 & 111,566 \\
Ant & FARL & 1,825.735 & 75.000 & \multicolumn{2}{c}{$\approx 0.2$M total (estimated)} \\
 & Eureka & 400.000 & 25.000 & 47,112 & 104,420 \\
 & LiMEN & 900.000 & 0 & 376,478 & 52,042 \\
Door & FARL & 9.100 & 0 & 795,008 & 78,751 \\
 & Eureka & 82.510 & 0 & 4,252 & 11,224 \\
 & LiMEN & 195.000 & 0 & 512,873 & 64,043 \\
Hand ST & FARL & 408.447 & 0 & 2,328,957 & 229,031 \\
 & Eureka & 80.249 & 30.147 & 108,732 & 175,236 \\
 & LiMEN & 31.000 & 24.041 & 536,643 & 70,776 \\
 & Native PPO & 0 & 15.000 & 0 & 0 \\
 & Native SAC & 0 & 15.000 & 0 & 0 \\
 & Native TD3 & 0 & 15.000 & 0 & 0 \\
\bottomrule
\end{tabular}
\end{table}

\paragraph{Ant token estimate.}
The retained usage JSONL contains one late 1,150-token request; launcher and
resume logs preserve 40 earlier successful requests without usage fields.
We assign these events to their surrounding module and estimate input plus
output tokens using module-specific mean usage from the Ant EA and DoorKey
EW archives. Three adapter calls use the mean task-construction cost.
The resulting estimate rounds to $0.2$M tokens; the artifact package records
all assignments and calibration values. Deterministic per-trial diagnoses
incur no LLM token charge.

\paragraph{Hand search expenditure.}
The main table reports the complete Hand search through iteration 15,
including the preceding 1M candidates at iterations 0--10 and all ten 5M
configurations at iterations 11--15. Its 126 worker files contain 262 distinct
seed receipts: 381,664,696 learner steps plus 26,781,982 source-policy prefix
interactions, totaling 408,446,678 training interactions. The ten 5M
configurations account for 150M learner steps and 1,497,767 prefix interactions
within this total. The selected iteration~13 design uses 15M learner steps
across its three seeds, with no prefix interactions.

The 135 recorded LLM requests contain 2,328,957 measured input and 229,031
output tokens. One request lacks usage metadata; estimating it at 26,050
tokens from measured requests to the same module gives approximately 2.58M
tokens overall. The search ends at the prescribed iteration 15 with the
iteration~13 design still selected.

Eureka's five generation rounds record 283,968 total tokens. Subtracting
the 175,236 completion tokens from the totals gives 108,732 input tokens;
the log's individual prompt field is not the sum over all requests in a round.
Its 79 candidate-training logs and five selected-reward retraining logs
record 80,248,832 and 5,079,040 steps, respectively, including PPO rollout
rounding. LiMEN's controller records 607,419 tokens. Its program archive
contains 31 completed single-seed short-training runs at 1M; the selected
interface has a separate three-seed 3M retraining, totaling 40M nominal
steps. The reported 5M evaluations of the same selected designs add
25,067,520 steps across five Eureka seeds and 15,040,512
across three LiMEN seeds. These completed runs bring their total training
costs to 110.395M and 55.041M steps, including the earlier
confirmations. Fixed-design training makes no additional LLM calls.
The native PPO/SAC/TD3 controls each train three seeds to 5M;
their 1M checkpoints belong to the same trajectories and add no extra training.

\paragraph{Training and cohort accounting.}
FARL DoorKey includes 728 distinct 100k receipts, three designs on five
300k seeds, and five selected-design 1M runs. Ant includes 4,794 training
receipts in 1,554 completed worker results, plus five-seed 5M and 10M
cohorts. Its retained receipts sum to 1,900.735M steps. Eureka DoorKey/Ant
use 80 candidates plus five selected-reward runs; LiMEN uses 30 candidates.
The selected LiMEN Ant cohort is already included in its search total.
FARL Door records 182 trial--seed receipts of 50k steps. Eureka Door includes
80.085M steps in eight completed candidates and 2.425M in two interrupted
candidates at their last log entries. LiMEN Door's completed 30-round process
allocates $30\times2$M short training plus $27\times5$M full training.
The earlier interrupted process in the same log is kept separately.
Search counts exclude miniRL sensing, separate evaluation rollouts, and
unrecorded interactions; mirrored logs and repeated receipts are deduplicated.

The available EW logs identify gateway model alias \texttt{gpt-5.4}, with one
FARL DoorKey learner-selection call to \texttt{gpt-5.4-mini}. Aliases do not
pin immutable upstream snapshots. FARL records requested temperatures
0--0.4, Eureka 1.0, and LiMEN 0.7; the GPT-5 request paths omit temperature
and \texttt{top\_p}, so those configured values are not effective sampling
temperatures. LiMEN's completion cap is 16,384; Eureka sets no explicit cap.
FARL's reasoning-model request builder enforces at least 16,000 completion
tokens, but historical per-call caps are not fully retained. Preserved prompts,
responses, generated source, and constructor receipts support reconstruction
at the available level of detail.

\subsection{Artifact Package}
\begin{table}[!t]
\centering\footnotesize
\caption{Reconstructing the paper's mechanisms and measurements.
The records distinguish a proposed design, its executed configuration,
and the evidence used to report its outcome.}
\label{tab:v12-artifact-guide}
\setlength{\tabcolsep}{4pt}
\begin{tabular}{@{}>{\raggedright\arraybackslash}p{.20\textwidth}
>{\raggedright\arraybackslash}p{.37\textwidth}
>{\raggedright\arraybackslash}p{.36\textwidth}@{}}
\toprule
Paper object & Reconstruction materials & Check \\
\midrule
Task and interface & Environment/dependency identities, field meanings,
adapter and evaluator definitions & Dynamics, input permissions, and success
criteria agree with the declared task. \\
Co-designed recipe & Observation/reward source, learner constructor,
hyperparameters, exploration profile & The reported candidate is one complete
executable combination. \\
Diagnostic test & Generation context, reference/treatment artifacts,
prediction, bound arms, and outcomes & A planned intervention can be
distinguished from an executed comparison. \\
Preserved alternatives & Component identities, family keys, selection and
retention records, actual training depths & Coverage and continued training
refer to actual designs, not only proposed pools. \\
Reported outcome & Fixed-design seeds, checkpoint identity, evaluation
episodes, and aggregation script & Equal-seed means and SD reproduce the
tables without candidate or checkpoint splicing. \\
Figures and controls & Plotting scripts, frozen result registries,
native-control checkpoint evaluations & All plotted cohorts and axis units
are recoverable from their recorded observations. \\
\bottomrule
\end{tabular}
\end{table}

The artifact package contains task definitions, dependency versions,
generated programs, available training configurations, seeds, and
per-seed evaluation results. Checkpoint identities are retained where available;
model weights are not required to regenerate the figures. Result files and plotting scripts reproduce
the reported comparisons, including all 121 qualifying Ant designs.
Hand result records cover the earlier search and all ten configurations from
the completed 5M iterations 11--15, including thirty per-seed evaluations and the
completed-search receipt. The plotted learning curves cover nine configurations
and twenty-seven episode logs; the complete result table additionally includes
the final raw-input SAC configuration. The representation experiment records
its prediction, result, and memory update.
Language models propose programs and configurations; reported measurements
come from executed experiments or identified external records. The authors
are responsible for the analyses and conclusions.

% Broader connections relocated from the V12 main-text Related Work section.
% Closest contribution-level comparisons remain in Section 2.
\section{Extended Comparison to Prior Work}\label{app:prior-work}
Section~\ref{sec:related} compares the most closely related interface-search
and diagnostic agents. Here we discuss connections to AutoRL, experimental
search, and exploration.

\subsection{Modular AutoRL and Executable Program Search}
AutoRL encompasses representation, algorithm, architecture, hyperparameter,
reward, and hybrid optimization~\citep{autorl_survey2022}; ARLO organizes
such choices into modular pipelines~\citep{arlo2023}. Eureka and Text2Reward
demonstrate executable reward generation~\citep{eureka2024,text2reward2024}.
LiMEN extends interface search to observations and rewards while preserving
diverse candidates~\citep{limen2026}. FARL jointly revises interface programs,
learners, and training settings. Diagnosis uses the resulting behavior to
decide which component or combination to change next.

\subsection{Diversity, Training Fidelity, and Exploration Frontiers}
MAP-Elites preserves diverse solutions~\citep{mouret2015mapelites}; population-based training (PBT)
adapts training populations~\citep{jaderberg2017pbt}; Hyperband and asynchronous successive halving (ASHA)
allocate fidelity through successive halving~\citep{li2018hyperband,li2020asha}.
FARL uses a successive-halving-style budget ladder, but also reserves training
for required comparisons, designs that are still improving, and alternatives
that short runs cannot yet distinguish. Diagnosis can therefore affect which
candidates receive more training even when their current scores are low.

Go-Explore returns to promising states before exploring further, while
reward machines represent task progress~\citep{ecoffet2021goexplore,
icarte2022rewardmachines}. FARL uses task milestones to identify where
progress stops and where an existing policy could help a new learner begin
exploration. Final evaluation starts from the original reset distribution
with assistance disabled
(Appendix~\ref{app:frontier-exploration}).

\subsection{Behavioral Diagnosis and Experimental Memory}
RDA uses visual subtask analysis to revise subtask specifications and reward
code under a fixed learning procedure~\citep{rda2026}. Prompt-to-Policy
combines code/video judgment, training-dynamics analysis, reward and
hyperparameter revision, and lessons from prior iterations
\citep{prompt2policy2026}. FARL uses numeric trajectories and available
predicates, and records each explanation with the tested design and conditions.

ReAct interleaves reasoning and environment actions~\citep{react2023};
Reflexion uses verbal feedback and episodic memory to improve subsequent
attempts~\citep{reflexion2023}. FunSearch, ReEvo, and MLAgentBench likewise
connect execution to search or accumulated feedback
\citep{funsearch2023,reevo2024,huang2024mlagentbench}. In RL, the same design
can behave differently across training seeds, initializations, and budgets.
FARL records these conditions when it tests an explanation. Its memory keeps
proposed explanations separate from experimental findings, allowing later
search to revise components and allocate training using the appropriate
evidence.

ENPIRE enables coding agents to revise policy-training code through real-robot
experiments after constructing fixed reset and verification
interfaces~\citep{enpire2026}. Agentic AutoResearch for Space Autonomy searches
a predefined configuration space and subjects its selected configuration to
an offline audit using reseeded evaluation and leave-one-out
pruning~\citep{spaceautoresearch2026}. These works share the goal of autonomous
experimentation; FARL focuses on retaining interface--solver alternatives
for diagnostic comparisons and using their outcomes to guide subsequent
design and training allocation.

\section{Native and Discovered Learning Designs}\label{app:design-comparison}
This section compares what the benchmark supplies with what each method
actually selects. We use the six design axes from Section~\ref{sec:problem}:
observations $\phi$, reward $r$, learner $L$, training parameters $\lambda$,
exploration $e$, and policy training budget $b$. The environment defines
observations, rewards, actions, and termination. A learning algorithm,
optimizer, and training budget must be chosen for the experiment.
``Native control'' below pairs the benchmark interface and reward with
the specified fixed learner.
The comparisons describe complete designs. They do not attribute a performance
difference to one component unless that component was tested while the others
were held fixed.

\subsection{What the Native Benchmarks Provide}\label{app:native-designs}
\begin{longtable}{@{}>{\raggedright\arraybackslash}p{1.4cm}>{\raggedright\arraybackslash}p{5.1cm}>{\raggedright\arraybackslash}p{5.7cm}@{}}
\caption{Native observations and expert-designed reward terms. The public
wrappers used in our experiments are specified separately below.}
\label{tab:native-designs}\\
\toprule Task & Native observation & Native reward and episode horizon \\
\midrule\endfirsthead
\toprule Task & Native observation & Native reward and episode horizon \\
\midrule\endhead
DoorKey & Symbolic $7\times7\times3$ egocentric image (object, color, state),
direction, and a textual mission. The numeric EW wrapper uses 147 normalized
image entries plus direction, giving 148 inputs. & Zero until reaching the
exit, then $1-0.9t/T$. The $5\times5$ task has $T=250$ steps. Pickup and door
opening receive no separate native reward. \\
\addlinespace
Ant & 27 coordinates: torso height, quaternion, eight joint angles, three
root linear velocities, three angular velocities, and eight joint velocities.
Global root $x,y$ are excluded. Contact-force features are disabled. & Forward
velocity plus a survival term, minus $0.5\|a\|_2^2$; contact cost is inactive
in this configuration. Episodes end on unhealthy state or at 1,000 steps.
The 27D vector is already the benchmark's chosen representation; EW retains
it while withholding native reward and reward metadata. \\
\addlinespace
Door & 39 coordinates: two 18D frames of hand/gripper and object state,
followed by the 3D goal. This includes supplied object and goal geometry,
even when native shaping is withheld. & Meta-World's dense reward combines
grasp/readiness shaping with door-opening progress, using gripper effort,
hand--handle geometry, and door angle; success sets reward to 10 in the v2
reward implementation. The experimental horizon is 500 steps. \\
\addlinespace
Hand & 39 coordinates: 30 hand positions and three 3D relative vectors
(palm--object, palm--target, object--target). The experiment's 84D ST wrapper
adds 36 velocities and absolute palm, object, and target positions. & Dense
reward is $-0.1d_{po}+\mathbb{1}_{h>0.04}(1-0.5d_{pt}-0.5d_{ot})
+10\mathbb{1}_{d_{ot}<0.1}+20\mathbb{1}_{d_{ot}<0.05}$.
Here $h$ is object height and $d$ denotes Euclidean distance. Episodes last
200 steps. Evaluation additionally requires five consecutive successful steps. \\
\bottomrule
\end{longtable}

Native Hand controls retain 39D inputs and train PPO, SAC, or TD3 to 5M;
their complete settings are in Table~\ref{tab:native-config}. Native Ant uses
27D inputs and PPO, with the settings and 1M/5M/10M evaluations in
Appendix~\ref{app:native-ant}. No native Door control is reported. DoorKey's
native completion rate is a supplied reference, for which a full learner
configuration and seed-level training archive were not recovered; it is not
used as a controlled comparison of training efficiency.

\subsection{DoorKey: Raw Inputs Can Suffice with a Suitable Solver}\label{app:design-doorkey}
\begin{table}[!ht]
\centering\footnotesize
\caption{Selected DoorKey EW designs. All reported policies use 1M steps.
Exact programs are included in the code supplement.}
\label{tab:design-doorkey}
\begin{tabular}{@{}>{\raggedright\arraybackslash}p{1.2cm}>{\raggedright\arraybackslash}p{3.55cm}>{\raggedright\arraybackslash}p{3.55cm}>{\raggedright\arraybackslash}p{3.55cm}@{}}
\toprule Axis & Eureka & LiMEN & FARL \\
\midrule
$\phi$ & Unchanged 148D public input. & 188D: public input plus visible-object
counts, nearest-object offsets, front-cell flags, time, and inferred key possession.
& Unchanged 148D public input. \\
$r$ & Key pickup/approach, door opening and doorway crossing, goal approach
and an exit proxy, with small action/time penalties. & Visible key/door/goal
changes and geometry-based shaping computed alongside the expanded observation.
& Changes in soft counts of visible key, door, and goal categories, conditioned
on pickup/toggle/forward actions; step penalty $0.0015$. \\
$L,\lambda$ & Fixed PPO; settings in Table~\ref{tab:baseline-solvers}.
& Fixed PPO; settings in Table~\ref{tab:baseline-solvers}.
& A2C with selected learning rate, discount, entropy and GAE parameters
(Table~\ref{tab:doorkey-config}). \\
$e$ & PPO policy sampling. & PPO policy sampling with entropy coefficient 0.01.
& Policy sampling; the selected coverage profile adds no intrinsic reward
or A2C entropy floor. \\
$b$ & $5\times1$M selected-reward runs. & One 1M selected run; no candidate
passes the full-training threshold. & $5\times1$M evaluation runs, following
100k screening and five-seed 300k comparisons. \\
\bottomrule
\end{tabular}
\end{table}

The designs differ in more than reward length or observation dimension.
Eureka explicitly encourages moving through the doorway and approaching
the exit, yet its selected policy stops short of task completion. LiMEN
constructs a larger representation but also retains PPO. FARL selects a
simpler public-input design with A2C. The controlled comparison in
Table~\ref{tab:doorkey-config} is the evidence that the solver configuration
matters: the same observation and reward reach 100\% completion with the
selected A2C configuration and 20\% with the compared PPO configuration.

Under EA, the public wrapper instead exposes the 83D full-grid state.
FARL's selected PPO design expands it to 115D and uses an event-based reward:
3 for first key acquisition, 6 for subsequent door opening, and 12 for
terminal success after those events, with a 0.5 penalty for dropping the
key before opening the door. Episode memory prevents repeated collection
of the first two bonuses. This EA program reads the wrapper's key-possession,
door-open, and terminal-success fields. Its information access therefore
differs from EW's visibility-based program: EA uses directly available task
events, while EW must work from the partial observation.
The retained selection receipt verifies 115D inputs, PPO, three seeds and
100k steps, but does not retain the selected trial's complete optimizer
configuration. The artifact package identifies this missing record explicitly.

Eureka EA retains the 83D input and fixed PPO, combining native reward with
key/door/goal geometry and a strong exact-goal-occupancy bonus. Its reported
cohort trains for 500k steps. LiMEN EA creates 273 features from the full
grid, object-relative coordinates, direction, front-cell contents and task
phase, while its selected reward is exactly the native reward. It uses fixed
PPO and three 2M runs. These programs make the role of expert information
concrete: the EA designs can directly use state and event information that
EW must work without.

\subsection{Ant: Different Rewards, Representations, and Learners}\label{app:design-ant}
\begin{table}[!ht]
\centering\footnotesize
\caption{Selected Ant EW designs. Native input and reward are in
Table~\ref{tab:native-designs}; fixed baseline solvers are in
Table~\ref{tab:baseline-solvers}.}
\label{tab:design-ant}
\begin{tabular}{@{}>{\raggedright\arraybackslash}p{1.2cm}>{\raggedright\arraybackslash}p{3.55cm}>{\raggedright\arraybackslash}p{3.55cm}>{\raggedright\arraybackslash}p{3.55cm}@{}}
\toprule Axis & Eureka & LiMEN & FARL \\
\midrule
$\phi$ & Unchanged 27D input. & 80D: orientation axes, bounded joint/velocity
features, pairwise joint coordination, summaries, and time. & 57D: all 27
public coordinates plus 30 pairwise coordinate differences. \\
$r$ & Forward velocity gated by height/uprightness, small posture bonuses,
and lateral, vertical, angular, joint-motion and action penalties.
& Forward velocity with posture-dependent scaling, small upright/facing
bonuses and motion/action penalties. & $0.8v_x+0.12\operatorname{Var}(a)
+0.08|\bar a_{0:4}-\bar a_{4:8}|$, minus coordinate-norm penalties. \\
$L,\lambda$ & Fixed PPO. & Fixed PPO. & SAC; $256,256$ network,
learning rate $2.5981\times10^{-4}$, $\gamma=0.991472$, $\tau=0.007961$. \\
$e$ & PPO policy sampling. & PPO policy sampling; entropy 0.01.
& Automatic SAC entropy initialized at 0.2; 10k random-action warmup. \\
$b$ & $5\times5$M reported runs. & $3\times10$M reported runs.
& Five runs per budget at 1M, 5M, and 10M; search first tests designs at
5k, 50k, 300k, and 1M. \\
\bottomrule
\end{tabular}
\end{table}

The selected EW reward encourages forward velocity and variation across
actuator commands. Its two action terms measure variance across all eight
action coordinates and the difference between the means of the first and
last four coordinates. The exact program also specifies the coordinate-norm
penalties and numerical bounds.

These design differences help explain what FARL searches, but the strongest
performance evidence remains the complete configurations and their training
cohorts. The 121-design comparison shows large variation even within SAC;
it does not establish that the action-diversity term alone causes the gain.

Under EA, FARL keeps the native 27D input and constructs the speed/control-cost
reward given in Appendix~\ref{app:ant-designs}. The saved SAC model uses
$256,256,128$ hidden layers, learning rate $3\times10^{-4}$,
$\gamma=0.990520$, $\tau=0.005448$, automatic entropy, and 20k warmup.
These are the effective settings recovered from the saved model.
Eureka EA retains 27D inputs and PPO; its reward adds linear and
exponential forward-speed terms to native reward, with lateral-drift and
unhealthy-state penalties. LiMEN's supplied EA program uses 30 features
(scaled native coordinates and three motion summaries) and PPO, with
forward velocity, a falling penalty, angular/joint-motion penalties and a
healthy-height bonus. Their reported policy budgets are 10M, 5M and 10M
for FARL, Eureka and LiMEN, respectively.

\subsection{Door: Several Different Designs Reach the Ceiling}\label{app:design-door}
\begin{table}[!ht]
\centering\footnotesize
\caption{Door EW designs. Eureka's reported 75/75 outcomes cover three
generated rewards, so all three programs are preserved.}
\label{tab:design-door}
\begin{tabular}{@{}>{\raggedright\arraybackslash}p{1.2cm}>{\raggedright\arraybackslash}p{3.55cm}>{\raggedright\arraybackslash}p{3.55cm}>{\raggedright\arraybackslash}p{3.55cm}@{}}
\toprule Axis & Eureka & LiMEN & FARL \\
\midrule
$\phi$ & Unchanged 39D input. & 121D: public-state transformations, hand/object/
goal relative vectors, opening-direction projections, alignment features,
state summaries and time. & 47D: 39 public coordinates plus eight generated
distance, offset, magnitude and configuration features. \\
$r$ & Three rewards combine reaching and opening progress, proximity bonuses
and action penalties. Two additionally gate opening or motion by proximity
to the handle. & Reaching progress, proximity-gated and alignment-weighted
opening progress, goal proximity, relative progress and a small action penalty.
& Negative hand--goal, object--goal and hand--handle distances; vertical-offset
and gripper terms; mean squared-action penalty. \\
$L,\lambda$ & Fixed PPO. & Fixed PPO. & SAC; $256,256$ network,
learning rate $4.0943\times10^{-4}$, $\gamma=0.981102$, $\tau=0.009037$. \\
$e$ & PPO policy sampling. & PPO policy sampling; entropy 0.01.
& Automatic SAC entropy; 10k warmup. \\
$b$ & Three designs, each 10M. & Three seeds, each 5M. & Two seeds, each 50k. \\
\bottomrule
\end{tabular}
\end{table}

All three methods complete this task. FARL reaches that ceiling with a
shorter trained policy, but the comparison does not identify one uniquely
correct reward or representation. It provides a useful contrast to Hand:
Door admits several effective geometric designs, whereas reaching and
lifting the object in Hand do not by themselves produce reliable transport.

\subsection{Hand: Preserve Contact While Improving the Training Configuration}\label{app:design-hand}
\begin{table}[!ht]
\centering\footnotesize
\caption{Selected Hand ST designs. Native controls use 39D inputs; all three
automated methods receive the 84D public wrapper. Policy budgets refer to
completed cohorts.}
\label{tab:design-hand}
\begin{tabular}{@{}>{\raggedright\arraybackslash}p{1.2cm}>{\raggedright\arraybackslash}p{3.55cm}>{\raggedright\arraybackslash}p{3.55cm}>{\raggedright\arraybackslash}p{3.55cm}@{}}
\toprule Axis & Eureka & LiMEN & FARL \\
\midrule
$\phi$ & Unchanged 84D public input. & 396D: native/public inputs, scaled joint
and velocity features, relative vectors in several frames, distances,
alignments, geometry summaries, time and native feedback.
& 111D: 84 public inputs, 20 other derived features and seven selected
relational features. \\
$r$ & Native reward plus smooth reach, lift, controlled-lift, target-proximity,
transport and late-placement terms, with a small action penalty.
& Half native reward plus reach/lift/transport changes normalized by scene
scale, contact-weighted progress and state-quality terms.
& Half native reward plus bonuses for maintaining control after lifting
and bringing the lifted object near target; penalty for losing contact. \\
$L,\lambda$ & Fixed PPO. & Fixed PPO. & SAC; $384,384,256$ network,
learning rate $1.2391\times10^{-4}$, $\gamma=0.987258$, $\tau=0.009802$. \\
$e$ & PPO policy sampling. & PPO policy sampling; entropy 0.01.
& Automatic entropy, target $-30$, and 1M random-action warmup. \\
$b$ & Five seeds, each 5M. & Three seeds, each 5M. & Three seeds, each 5M. \\
\bottomrule
\end{tabular}
\end{table}

Let $d_{po}$ and $d_{ot}$ denote next-state palm--object and object--target
distance, $h$ object height, and $r_N$ native reward. FARL's selected program is
\[
r=0.5r_N+3\mathbb{1}_{h>0.04}\mathbb{1}_{d_{po}<0.10}
 +2\mathbb{1}_{h>0.04}(0.15-d_{ot})_+
 -2.5\mathbb{1}_{h>0.04}(d_{po}-0.12)_+.
\]
It keeps native feedback during approach and adds explicit incentives for
controlled transport after lifting. Eureka and LiMEN also contain transport
shaping, so the result cannot be explained by claiming that only FARL rewards
the later stage. The distinguishing evidence is the search over the whole
configuration. The same generated reward and 111D representation achieve
only 4.17\% at iteration 8; longer training and revised SAC settings eventually
raise the selected design to \handSuccess\%. The matched 104D/111D test and
the subsequent solver revisions are detailed in Appendix~\ref{app:hand-breakthrough}.

\subsection{Learner Settings and Exact Program Records}\label{app:design-settings}
\begin{table}[!ht]
\centering\footnotesize
\caption{Fixed automated-baseline PPO settings. These are experiment
settings, not hyperparameters supplied by the environments.}
\label{tab:baseline-solvers}
\begin{tabular}{@{}lcc@{}}
\toprule Setting & Eureka & LiMEN \\
\midrule
Policy / value hidden layers & $64,64$ / $64,64$ & $64,64$ / $64,64$ \\
Activation & Tanh & Tanh \\
Learning rate & $3\times10^{-4}$ & $3\times10^{-4}$ \\
Discount / GAE & $0.99/0.95$ & $0.99/0.95$ \\
Clipping / value coefficient & $0.2/0.5$ & $0.2/0.5$ \\
Batch size / optimization epochs & $64/10$ & $256/10$ \\
Entropy coefficient & 0 & 0.01 \\
Gradient norm limit & 0.5 & 0.5 (Ant: 1.0) \\
Replay / random-action warmup & None & None \\
\bottomrule
\end{tabular}
\end{table}

Eureka uses eight parallel environments and 2,048 steps per environment per
rollout. LiMEN uses eight environments with 2,048 steps each for Hand,
32 with 2,048 for Door, 48 with 1,365 for Ant, and eight with 8,192 for
DoorKey. These are the SB3 rollout lengths derived from the task configurations.
FARL's selected SAC designs use a replay capacity of $10^6$, batch size 256,
and one gradient update per environment step. The effective parameters
listed above are taken from saved models where available, rather than
assuming that every field in an LLM proposal was accepted by the learner.

The supplement pairs each retained program with its task, method, candidate
identity, observation dimension, training budget, and available constructor
or selection records. It includes executable baseline reward/interface
programs and FARL observation/reward specifications. This makes it possible
to inspect what was designed, as well as compare the final scores.

\end{document}